\documentclass[preprint,12pt]{elsarticle}

\usepackage{amssymb}
\usepackage{amsmath}
\usepackage{makeidx}
\usepackage{latexsym}
\usepackage{paralist}
\usepackage{empheq} 
\usepackage{verbatim}
\usepackage{url}
\usepackage{stmaryrd}
\usepackage{enumitem}
\usepackage{bussproofs}
\usepackage{boxedminipage}
\usepackage[switch]{lineno}
\newtheorem{theorem}{Theorem}

\newtheorem{lemma}{Lemma}
\newtheorem{example}{Example}
\newtheorem{proof}{Proof}
\newtheorem{definition}{Definition}
 \usepackage{multicol}
 
 \usepackage{wrapfig}
\usepackage{lscape}
\usepackage{rotating}
\usepackage{graphicx}
\usepackage{caption}
\usepackage{amsmath}
\usepackage{tcolorbox}
\usepackage[export]{adjustbox}

{

	\newcommand{\typearrow}{\shortrightarrow}
	\newcommand{\itype}{i}
	
	\newcommand{\proptype}{\tau}
	\newcommand\detailedproof[1]{}
	
	\newcommand{\opf}{\mathrm{opt}_{f}}
\journal{XXX}
\begin{document}

\begin{frontmatter}



\title{New Algebraic Normative Theories for Ethical and Legal Reasoning in the LogiKEy Framework }

%\\
%farjami110@gmail.com
%\\
%University of Luxembourg
%}

\author{Ali Farjami
\\
farjami110@gmail.com,
University of Luxembourg
}
%\affiliation{
%organization={Department of Computer Science, University of %Luxembourg},
%addressline={farjami110@gmail.com}, 
%country={Luxembourg}}
%\hspace{-.5cm}
\begin{abstract}
 In order to design and engineer ethical and legal reasoners and responsible systems, Benzm\"{u}ller, Parent and van der Torre introduced the LogiKEy methodology, based on the semantical embedding of deontic logics into classic higher-order logic.  This article considerably extends the LogiKEy deontic logics and dataset using an algebraic approach, and develops a theory of input/output operations for normative reasoning on top of Boolean algebras.
 %We characterize a class of proof systems over Boolean algebras for a set of explicitly given conditional norms.
 %We develop theory of input/output operations for normative reasoning on top of Boolean algebras.
%Input/output logic investigates abstract study of normative concepts in terms of procedures yielding outputs for inputs. In this paper, we provide an algebraic characterization for these input/output operations. These operations are implemented in Isabelle/HOL. Moreover, we show that how  we can integrate normative reasoning with preferences. 
\end{abstract}


%\item LogiKEy Framework
%\item Normative Reasoning


\end{frontmatter}


\section{Introduction}
\label{}
Benzm\"{u}ller, Parent and van der Torre~\cite{J48} introduced the LogiKEy framework  for the formalization and automation of new ethical reasoners, normative theories and deontic logics. The LogiKEy framework uses higher-order logic (HOL) as a metalogic to embed other logics. A logic embedded in HOL can thus be tracked by automated theorem provers (ATP), interactive automated provers (ITP) and HOL model finders.  The LogiKEy methodology allows a user to simultaneously combine and experiment with underlying logics (and their combinations), ethico-legal domain theories, and concrete examples.

Earlier work presented semantical embedding of two traditions in deontic logic in the LogiKEy framework, namely  \AA{}qvist’s dyadic deontic logic {\bf E}~\cite{J45} and Makinson and van der Torre’s input/output (I/O) logic~\cite{J46}.  Subsequent work provided the Isabelle/HOL dataset for the LogiKEy workbench~\cite{J53}.
This article considerably extends the LogiKEy deontic logics and dataset using an algebraic approach. In particular, it extends the theory of input/output operations~\cite{J46} and corresponding proof systems on top of Boolean algebras and, more generally, abstract logics~\cite{Jansana}.

 The article is structured as follows: Section~\ref{sec: BOP} and \ref{sec: BOO} provide the soundness and completeness results of I/O operations for deriving permissions and obligations on top of Boolean algebras. Section~\ref{sec: Ab} shows how I/O operations can be generalized over any abstract logic. Section~\ref{sec : SNP} integrates a conditional theory into input/output logic.  Section~\ref{sec: BIO} introduces semantical embedding of I/O logic into HOL, including soundness and completeness (faithfulness).  Section~\ref{sec: RW} discusses related work, and Section~\ref{sec: CN} concludes the article. Appendix A shows how the semantical embedding described in Section~\ref{sec: BIO} is implemented in the Isabelle/HOL proof assistant. Some experiments are provided to show that this logic implementation enables interactive and automated reasoning. All the proofs are in the appendices. Appendix B provides proofs relating to Sections~\ref{sec: BOP} and \ref{sec: BOO}, Appendix C proofs relating to Section~\ref{sec : SNP} and Appendix D proofs relating to Section~\ref{sec: BIO}.
 %Section~\ref{sec: FR} discusses future work and
 %\newpagesec: FR
 %$ \top  $ denotes truth in all possible worlds and $ \bot $ stands for $ \neg \top $

 \section{Permissive norms: input/output operations} \label{sec: BOP}
 Deontic logic investigates logical relations among normative concepts~\cite{sep-logic-deontic}. Input/output (I/O) logic was initially introduced by Makinson and van der Torre~\cite{Makinson00} to study  conditional norms viewed as relations between logical formulas~\cite{Makinson00}. In this setting, the meaning of normative concepts is given in terms of a set of procedures yielding outputs for inputs. Suppose that $ N^{O} $ denotes a set of obligatory norms and $ N^{P} $ a set of permissive norms. The formula $(a,x) \in N^{O} $ means ``given $ a $, it is obligatory that $ x $'', while the formula $(a,x) \in N^{P} $ means ``given $ a $, it is permitted that $ x $.'' The formula $x \in out(N^{P}, A)$ means ``given  normative system $N^{P}$ and input set $ A $ (state of affairs), $x$ (permission) is in the output''. The output operations resemble inferences, where inputs need not be included among outputs, and outputs need not be reusable as inputs~\cite{Makinson00}. The proof system of an I/O logic is specified via a number of derivation rules acting on pairs $(a,x)$ of formulas. Given a set $N$ of pairs, $(a,x)\in deriv_i (N)$ is written to say that $(a,x)$ can be derived from $N$ using these rules. The term ``input/output logic'' is used broadly to refer to a family of related systems such as \textit{simple-minded}, \textit{basic}, and \textit{reusable} systems~\cite{parent2018introduction,Makinson00}.  This section uses similar terminology, and introduces some input/output systems for deriving permissions on top of Boolean algebras. Each derivation system is closed under a set of rules, including for instance the weakening of the output (WO) rule or the strengthening of the input (SI) rule.   A bottom-up approach is used to characterize different derivation systems. The AND rule, for the output, is absent in the derivation systems presented in this section.
 %Systems are defined that are closed in order to allow for example, the weakening of the output (WO) rule or the strengthening of the input (SI) rule.
 %\pagebreak
 \begin{definition} [Boolean algebra] A structure $\mathcal{B} = \langle B, \wedge, \vee, \neg, 0, 1 \rangle $ is a Boolean algebra if and only if it satisfies the following identities:\footnote{An equation $ t\approx t^{'} $ holds in an algebra $\mathcal{A}$ if its universal closure $ \forall x_{0}...x_{n} t\approx t^{'} $ is a sentence that is true in $\mathcal{A}$.}
 \vspace{.5cm}
 	\begin{compactitem}
 		\item  $ x \vee y \approx y \vee x $,  $ x\wedge y \approx y \wedge x $
 		\item  $ x \vee ( y \vee z) \approx (x \vee y) \vee z$, $ x \wedge (y \wedge z) \approx (x \wedge y ) \wedge z$
 		\item  $ x \vee 0\approx x $,  $ x \wedge 1 \approx x $
 		\item  $ x \vee \neg x  \approx 1 $,  $ x\wedge \neg x \approx 0 $
 		\item   $ x \vee (y\wedge z)\approx (x \vee y) \wedge (x \vee z) $,  
 		$ x \wedge (y \vee z) \approx (x \wedge y) \vee (x \wedge z)$
 	\end{compactitem}
%	The elements of a Boolean algebra are ordered as $ a \leq b $ iff $  a \wedge b = a$.
\end{definition}
%\vspace{.5cm}
 %%%%%%%%%%%%%%%%%%%%%%%%%%%%%%
 \begin{definition} [Syntax]
 For a set of variables $ X $,  the set of Boolean terms is denoted over $ X $ by $ Ter(X) $ as follows:
 \begin{center}
 	$ Ter(X) = \bigcup_{n \in N} Ter_{n}(X)$
 \end{center}
where 
\begin{center}
	$ Ter_{0} (X) = X \cup \{0,1\}  $
	\\
	$ Ter_{n+1} (X) = Ter_{n} (X) \cup \{ a \wedge b, a\vee b, \neg a: a, b \in Ter_{n} (X) \} $.
\end{center}
Given a Boolean algebra $ \mathcal{B} $, the elements of  $ Ter(B) $ are ordered as $ a \leq b $ iff $  a \wedge b = a$.\footnote{The symbol ``$=$'' is used to express that both sides name the same object. The elements of the variable set ($ B $) that are represented by different letters are supposed to be independent in the algebra ($\mathcal{B}$) w.r.t $ \leq $. } Since $\leq$ is antisymmetric, $ a \leq b $ and $ b \leq a $ imply $ a=b $.
 
%%%%%%%%%%%%%%%%%%%%%%%
 \end{definition}
 
 \begin{definition}[Upward-closed set]
 	Given a Boolean algebra $ \mathcal{B} $, a set \linebreak $ A \subseteq Ter(B) $ is called upward-closed if it satisfies the following property:
 
 	\begin{center}
 		For all $x,y \in Ter(B)  $, if $ x  \leq y $ and $ x \in A $, then $ y\in A $.
 	\end{center}
 	The least upward-closed set that includes $ A $ by $ Up(A) $ is denoted. The $ Up $ operator satisfies the following properties:
 	\begin{compactitem}
 		\item $ A \subseteq Up(A) $ \hfill (Inclusion)
 		\item $ A \subseteq B \Rightarrow Up(A) \subseteq Up(B)$ \hfill (Monotony) 
 		\item $ Up(A) = Up(Up(A)) $ \hfill (Idempotence)
 	\end{compactitem}
 	An operator that satisfies these properties is called a closure operator. 
 \end{definition} 
 The ``$ Up $'' operator, for a given set $ A $, sees all the elements that are in a higher or equal position to the elements of $ A $ in terms of their usual ordering in lattices. Unlike the 
 \textit{propositional logic consequence relation} (``$Cn$'') operator, the ``$ Up $'' operator is not closed under conjunction so that we do not have $ a \wedge \neg a \in Up(a, \neg a) $. 
 %%%%%%%%%%%%%%%%%%%%%%%%%%%%%%%%%%%%%%%%

  \begin{definition}[Semantics] In input/output logic, the main semantic construct for normative propositions is the output operation, which represents the set of normative propositions related to a normative system $N$, regarding state of affairs $A$, namely $out(N,A)$.  
  Normative system $ N $ denotes a set of norms ($a$, $x$) in which the body and head are Boolean terms.
 Let $N(A) = \{ x \mid (a,x) \in N \text{ for some } a \in A \}$. In a Boolean algebra $ \mathcal{B} $ for $ X \subseteq Ter(B) $, the equation $ Eq(X) = \{x \in Ter(B) | \exists y \in X, x = y \}$ is defined.\footnote{ 
 			%	$ N^{*}(A) = \{x| (b,y) \, \text{for some } \, b \in A \, \text{such that} \, x=y, a=b \} $
 			Sometimes $Up(a,b,...)  (Eq(a,b,...))$ is written instead of $Up(\{a,b,...\})  (Eq(\{a,b,...\}))$ and $out(N, a)$  $(derive(N, a)) $ is written instead of $out(N, \{a\})$ $ (derive(N, \{a\})) $.}
 	Given a Boolean algebra $ \mathcal{B} $, a normative system $ N \subseteq Ter(B) \times Ter(B) $ and an input set $ A \subseteq Ter(B)$, I/O Boolean operations are defined as follows:
 	\begin{description}
 	    \item Zero Boolean I/O operation: 
 	    $$ out^{\mathcal{B}}_{0} (N, A) = Eq(N (Eq(A)))$$
 	    $$ out^{\mathcal{B}}_{R} (N, A)\linebreak = Eq(N (A)) \, \, \,  out^{\mathcal{B}}_{L} (N, A) = N (Eq(A))$$
 		\item Simple-I Boolean I/O operation:
 		   $$ out^{\mathcal{B}}_{I} (N, A) = Eq(N (Up(A)))$$
 		\item Simple-II Boolean I/O operation: 
 		$$ out^{\mathcal{B}}_{II} (N, A) = Up(N(Eq(A)))$$
 		\item Simple-minded Boolean I/O operation:
 		$$ out^{\mathcal{B}}_{1} (N, A) = Up(N(Up(A)))$$
 		\item Basic Boolean I/O operation:\footnote{A set $ V $ is saturated in a Boolean algebra $ \mathcal{B} $ iff 
 	\begin{compactitem}
 		\item If $ a \in V $ and $ b \geq a $, then $ b \in V $;
 		\item If  $a \vee b \in V$, then $ a \in V $ or $ b \in V $. 
  	\end{compactitem}}
 		$$ out^{\mathcal{B}}_{2} (N, A) =  \bigcap \{ Up(N(V)), A \subseteq V, V \text{is saturated} \}$$
 		\item Reusable Boolean I/O operation:
 		$$ out^{\mathcal{B}}_{3} (N, A) =  \bigcap \{ Up(N(V)), A \subseteq V= Up(V) \supseteq N(V) \}$$
 	\end{description}
 	
 	Put $  out^{\mathcal{B}}_{i} (N) =  \{ (A, x) : x \in out^{\mathcal{B}}_{i} (N, A) \} $.
 \end{definition}
 
%A derivation of a pair (a, x) from N, given a setR of rules, is understood to be a tree with (a, x) at the root, each non-leaf node Consistent reusability related to its immediate parents by the inverse of a rule in R, and each leaf node an element of N.

We turn to the proof theory. A derivation of a pair $(a,x)$ from $N$, given a set $X$ of rules, is understood to be a tree with $(a,x)$ at the root, each non-leaf node related to its immediate parents by the inverse of a rule in $X$, and each leaf node an element of $N$.
 
  \begin{definition}[Proof system]
 	Given a Boolean algebra $ \mathcal{B} $ and a normative system $ N \subseteq Ter(B) \times Ter(B) $, it is defined that $(a,x) \in derive^{\mathcal{B}}_{i}(N) $ if and only if $ (a,x) $ is derivable from $ N $ using $EQI, EQO, SI, WO, OR, T$ as follows:\footnote{EQI  stands for equivalence of the input, EQO for equivalence of the output, and T for transitivity.} 
 	% SI   abbreviate ``strengthening of the input''.
 	%EQI  stands for equivalence of the input
 	%EQO stands for equivalence of the output
 	\begin{multicols}{3}
\hspace*{-.3in} 
	\begin{tabular}{l c}
		\hline
		\emph{derive}$^{\mathcal{B}}_i$ & Rules   \\
		\hline
		\emph{derive}$^{\mathcal{B}}_{R}$  & \{EQO\}   \\
		\emph{derive}$^{\mathcal{B}}_{L}$  & \{EQI\}   \\	
		\emph{derive}$^{\mathcal{B}}_{0}$  & \{EQI, EQO\}   \\
		\emph{derive}$^{\mathcal{B}}_{I}$  & \{SI, EQO\}   \\
		\emph{derive}$^{\mathcal{B}}_{II}$  & \{WO, EQI\}   \\
		\emph{derive}$^{\mathcal{B}}_{1}$  & \{SI, WO\}   \\
		\emph{derive}$^{\mathcal{B}}_{2}$  & \{SI, WO, OR\}   \\
		\emph{derive}$^{\mathcal{B}}_{3}$  & \{SI, WO, T\}  \\
		%\emph{deriv}$^{\mathcal{B}}_{4}$  &  \{SI, WO, OR, T\}\\
		\hline
	\end{tabular}   
	
   \vspace*{-.2in}
   \begin{prooftree}\hspace*{.1in}
   	\AxiomC{$(a,x)$} \AxiomC{$x = y$ }
   	\LeftLabel{EQO} \BinaryInfC{$(a,y)$}
   \end{prooftree}
   \begin{prooftree}\hspace*{.1in} 
   	\AxiomC{$(a,x)$} \AxiomC{$a = b$ }
   	\LeftLabel{EQI} \BinaryInfC{$(b,x)$}
   \end{prooftree}
   \begin{prooftree}\hspace*{.1in}
   	\AxiomC{$(a,x)$ } \AxiomC{$b\leq a$ }
   	\LeftLabel{SI} \BinaryInfC{$(b,x)$}
   \end{prooftree}

   \begin{prooftree}   
   	\AxiomC{$(a,x)$ } \AxiomC{$ (x, y)$}
   	\LeftLabel{T} \BinaryInfC{$(a,y)$}
   \end{prooftree}
   \begin{prooftree} 
   	\AxiomC{$(a,x)$ } \AxiomC{$ (b, x)$}
   	\LeftLabel{OR} \BinaryInfC{$(a \vee b,x)$}
   \end{prooftree}
   \begin{prooftree} 
   	\AxiomC{$(a,x)$} \AxiomC{$x\leq y$ }
   	\LeftLabel{WO} \BinaryInfC{$(a,y)$}
   \end{prooftree}
\end{multicols}
 	
 	Given a set of $ A \subseteq Ter(B) $, then $ (A,x) \in  derive^{\mathcal{B}}_{i}(N) $ whenever $ (a,x) \in derive^{\mathcal{B}}_{i}(N) $ for some $ a \in A$.\footnote{In the work of Makinson and van der Torre~\cite{Makinson00}, it is for some conjunction $ a $ of elements in $ A $.} Put $ derive^{\mathcal{B}}_{i}(N,A) = \{x : (A,x) \in derive^{\mathcal{B}}_{i}(N)\}$.
 \end{definition}
 
 \begin{theorem} [Soundness and completeness]\label{th:out}
 	$  out^{\mathcal{B}}_{i}(N) = derive^{\mathcal{B}}_{i}(N) $.  
 \end{theorem}

\begin{example}
	 For the conditionals $ N= \{(1, g), (g,t)\} $ and the input set $ A = \{\}$ we have $ out^{\mathcal{B}}_{I}(N,A) = \{\}$, and for the input set $ C = \{g\}$ we have $ out^{\mathcal{B}}_{II}(N,C) =  Up(t)  $.
\end{example}

% \begin{example}
%  For the conditionals $ N= \{(1, g), (g,t)\} $ and the input set $ A = \{\}$ we have $ out^{\mathcal{B}}_{II}(N,A) = \{\}$, and for the input set $ C = \{g\}$ we have $ out^{\mathcal{B}}_{II}(N,C) =  Up(t)  $.
% \end{example}

% \begin{example}
% For the conditionals  $N = \{(g, t), (\neg g, \neg t), (a, b) \} $ and the input set $ A= \{g, \neg g\} $ we have $ out^{\mathcal{B}}_{1}(N, A) = Up (t, \neg t )$.
% \end{example}

 \begin{example}
 	For the conditionals  $N = \{(1, g), (g, t), (\neg g, \neg t ), (a, b) \} $ and the input set $ A= \{ \neg g\} $ we have $ out^{\mathcal{B}}_{3}(N, A) = Up (g, t, \neg t )$.
 \end{example}

 \section{Obligatory norms: input/output operations} \label{sec: BOO}	
 This section adds the AND and cumulative transitivity (CT) rules to the derivation systems introduced, with the aim of rebuilding the derivation systems introduced by Makinson and van der Torre~\cite{Makinson00} for deriving obligations. 
 
 \begin{definition}[Proof system]\label{def:ob}
 	Given a Boolean algebra $ \mathcal{B} $ and a normative system $ N \subseteq Ter(B) \times Ter(B) $, it is defined that $(a,x) \in derive^{X}_{i}(N) $ if and only if $ (a,x) $ is derivable from $ N $ using $EQO, EQI, SI, WO, OR, AND, CT$ as follows:
 	%\hspace*{-.2in}

% \newpage	
 	\begin{multicols}{2}
 		%\vspace*{1.5in}
 		\begin{tabular}{ll}
 			\hline
 			\emph{derive}$^{X}_i$ & Rules   \\
 			\hline	
 			\emph{derive}$^{AND}_{II}$  & \{WO, EQI, AND\}   \\
 			\emph{derive}$^{AND}_{1}$  & \{SI, WO, AND\}   \\
 			\emph{derive}$^{AND}_{2}$  & \{SI, WO, OR, AND\}   \\
 			\emph{derive}$^{CT}_{I}$  & \{SI, EQO, CT\}   \\
 			\emph{derive}$^{CT}_{II}$  & \{WO, EQI, CT\}   \\
 			\emph{derive}$^{CT}_{1}$  & \{SI, WO, CT\}   \\
 			%\emph{deriv}$^{CT,AND}_{I}$  & \{SI, EQO, CT, AND  \} \\
 			\emph{derive}$^{CT,AND}_{1}$  & \{SI, WO, CT, AND\}   \\
 			\hline
 		\end{tabular}
 		%\vspace*{-.4in}
 		\begin{prooftree} 
 			\AxiomC{$(a,x)$ } \AxiomC{$ (a,y)$}
 			\LeftLabel{AND} \BinaryInfC{$(a, x\wedge y)$}
 		\end{prooftree} 
 		%\vspace*{-.4in}
 		\begin{prooftree}
 			\AxiomC{$(a,x)$ } \AxiomC{$ (a\wedge x,y)$}
 			\LeftLabel{CT} \BinaryInfC{$(a,y)$}
 		\end{prooftree}
 	\end{multicols}{}
 	
 	Given a set of $ A \subseteq Ter(B) $, $ (A,x) \in  derive^{X}_{i}(N) $ whenever $ (a,x) \in derive^{X}_{i}(N) $ for some $ a \in A$. Put $ derive^{X}_{i}(N,A) = \{x : (A,x) \in derive^{X}_{i}(N)  \}  $.
 \end{definition}
 
% With this understanding, the order of application of two derivation rules is often ‘reversible’. In some cases, we may simply permute the application of two successive rules, independently of the choice of the formulae to which they are applied. For example, any application of AND followed by SI may be replaced by one in which SI is followed by AND. In other cases,
%the order may be reversed, but with additional (and prior) use of a third rule – often SI and in one instance WO. Finally, there are some configurations for which no transformation appears to be available.

 Makinson and van der Torre~\cite{Makinson00} noticed that in some cases, the order of application of two derivation rules is \textit{reversible}. For instance, any application of AND followed by WO (SI) may be replaced by one in which WO (SI) is followed by AND. Based on this observation, new output operations are defined by,  
 for example, rearranging the derivation $(a,x)$ in the proof system $\{SI, WO, AND\}$ such that the AND rule applies only at the end. The $\{SI, WO\}$ system has been characterized as the \textit{simple-minded I/O operation}~$out^{\mathcal{B}}_{1}$. Now by using  (finite) iterations of AND on top of $out^{\mathcal{B}}_{1}$, a new output operation is defined that characterizes the proof system $\{SI, WO, AND\}$. Three kinds of such output operations are defined--- $ out^{AND}_{i} $, $ out^{CT}_{i} $, and  $ out^{CT,AND}_{i} $ ---that can characterize the proof systems introduced in Definition~\ref{def:ob}. Note that there are some non-reversible orders, such as the WO rule followed by OR rule, for which no transformation appears to be available.

 \begin{definition}[Semantics $ out^{AND}_{i} $]\label{outAND}
 	Given a Boolean algebra $ \mathcal{B} $, a normative system $ N \subseteq Ter(B) \times Ter(B) $ and an input set $ A \subseteq Ter(B)$, the AND operation is defined as follows: \[\begin{array}{ll}
 	out^{AND^{0}}_{i} (N, A)&= out^{\mathcal{B}}_{i} (N, A)
 	\\
 	out^{AND^{n+1}}_{i} (N, A)&= out^{AND^{n}}_{i} (N, A) \, \,     \cup 
 	\\
 	& \{  y \wedge z: y,z \, \in out^{AND^{n}}_{i} (N, \{a\}), \,  \, \, a \in A \}
 	\\
 	out^{AND}_{i} (N, A) &= \bigcup_{n\in N}  out^{AND^{n}}_{i} (N, A)
 	\end{array}
 	\]
 	Put $  out^{AND}_{i} (N) =  \{ (A, x) : x \in out^{AND}_{i} (N, A) \} $.
 \end{definition}

 \begin{definition}[Semantics $ out^{CT}_{i} $]
 	Given a Boolean algebra $ \mathcal{B} $, a normative system $ N \subseteq Ter(B) \times Ter(B) $ and an input set $ A \subseteq Ter(B)$, the CT operation is defined as follows:
 	\[
 	\begin{array}{ll}
 	out^{CT^{0}}_{i} (N, A)&= out^{\mathcal{B}}_{i} (N, A)
 	\\
 	out^{CT^{n+1}}_{i} (N, A)&= out^{CT^{n}}_{i} (N, A) \, \,     \cup 
 	\\
 	& \{x: y \in\, out^{CT^{n}}_{i} (N, \{a\}) \, \text{and} \, x \, \in out^{CT^{n}}_{i} (N, \{a \wedge y\}), \, \, \, a \in A \}   
 	\\
 	out^{CT}_{i} (N, A) &= \bigcup_{n\in N}  out^{CT^{n}}_{i} (N, A)
 	\end{array}
 	\]
 	Put $  out^{CT}_{i} (N) =  \{ (A, x) : x \in out^{CT}_{i} (N, A) \} $.
 \end{definition}

 \begin{definition}[Semantics $ out^{CT,AND}_{i} $]
 	Given a Boolean algebra $ \mathcal{B} $, a normative system $ N \subseteq Ter(B) \times Ter(B) $ and an input set $ A \subseteq Ter(B)$, the CT,AND operation is defined as follows:
 	\[
 	\begin{array}{ll}
 	out^{CT,AND^{0}}_{i} (N, A)&= out^{CT}_{i} (N, A)
 	\\
 	out^{CT, AND^{n+1}}_{i} (N, A)&= out^{CT, AND^{n}}_{i} (N, A) \, \,     \cup 
 	\\
 	& \{  y \wedge z: y,z \, \in out^{CT, AND^{n}}_{i} (N, \{a\}), \, \, \, a \in A \}
 	\\
 	out^{CT, AND}_{i} (N, A) &= \bigcup_{n\in N}  out^{CT, AND^{n}}_{i} (N, A)
 	\end{array}
 	\]
 	Put $  out^{CT, AND}_{i} (N) =  \{ (A, x) : x \in out^{CT, AND}_{i} (N, A) \} $.
 \end{definition}

 \begin{theorem}\label{ob : th}
 	Given a Boolean algebra $ \mathcal{B} $, for every normative system \linebreak $ N \subseteq Ter(B) \times Ter(B) $ we have $ out^{AND}_{i} (N) = derive^{AND}_{i}(N) $, $ i\in \{II, 1, 2\} $; $ out^{CT}_{i} (N) = derive^{CT}_{i}(N) $, $ i\in\{I,II,1\} $;  and $ out^{CT, AND}_{1} (N) =$ $ derive^{CT, AND}_{1}(N)$.
 \end{theorem}

 Similarly, it is possible to define the $out_{i}^{OR}(N)$ operation and characterize some other proof systems:
 
 \begin{center}
 	\begin{tabular}{ c c  }
 		\hline
 		\emph{derive}$^{X}_i$ & Rules   \\
 		\hline	
 		\emph{derive}$^{OR}_{I}$  & \{SI, EQO, OR\}   \\
 		\emph{derive}$^{CT,OR}_{I}$  & \{SI, EQO, CT, OR\}   \\
 		%\emph{deriv}$^{CT,OR}_{II}$  & \{WO, EQI, CT, OR\}   \\
 		\emph{derive}$^{CT, OR}_{1}$  & \{SI, WO, CT, OR\}   \\
 		%\emph{deriv}$^{CT,OR,AND}_{I}$  & \{SI, EQO, CT, OR, AND\} \\
 		\emph{derive}$^{CT,OR,AND}_{1}$  & \{SI, WO, CT, OR, AND\}   \\
 		\hline
 	\end{tabular}
 \end{center}
 \begin{multicols}{2}
  
 \end{multicols}
 
 Four systems, namely $  derive^{AND}_{1}$, $ derive^{AND}_{2}$ (or $ derive^{OR,AND}_{1}$ ), \linebreak  $ derive^{CT,AND}_{1}$ and  $ derive^{CT,OR,AND}_{1}$, are introduced by Makinson and van der Torre~\cite{Makinson00} for reasoning about obligatory norms.
 %%%%%%%%%%%%%%%%%%%%%%%%%%%%%%%%%%%%%%%%%%%%%%%%%%%%%%%
 %New subsection maybe remove
  
 %%%%%%%%%%%%%%%%%%%%%%%%%%%%%%%%%%%%%%%%%%%%%%%%%%%%%%%%%
 \section{I/O mechanism over abstract logics}\label{sec: Ab}
 An abstract logic~\cite{Jansana} is a pair $\mathcal{A} = \langle \mathcal{L} , C\rangle $ where $ \mathcal{L} = \langle L, ...\rangle $ is an algebra and $ C $ is a closure operator, defined on the power set of its universe,  that means that for all $ A, B \subseteq L $:
 \begin{compactitem}
 	\item $ A \subseteq C(A)  $
 	\item $ A \subseteq B \Rightarrow C(A) \subseteq C(B) $
 	\item $ C(A) = C (C (A)) $
 \end{compactitem}

 The elements of an abstract logic can be ordered as $ a \leq b $ if and only if $ b \in C(\{a\}) $.\footnote{$ a=b $ if and only if $ a \leq b $ and $ b \leq a$.} Without loss of generality, the algebra of formulas (or terms in the algebraic context) is used where $ \mathbf{Fm}(X)= \langle Fm(X), ...\rangle $ for a set of fixed variables $ X $. Similar to Boolean algebras, the $ Eq $ and $Up$ operators can be defined for $ A \subseteq Fm(X)$. 
 
 \begin{definition}[Semantics]
 	Given an abstract logic $\mathcal{A} = \langle \mathbf{Fm}(X), C\rangle $, a normative system $ N \subseteq Fm(X) \times Fm(X)$ and an input set $ A  \subseteq Fm(X)$, the I/O operations are defined as follows:
 	\begin{compactitem}
 		\item $ out^{\mathcal{A}}_{0} (N, A) =Eq(N(Eq(A)))$
 		\item $ out^{\mathcal{A}}_{I} (N, A) =Eq(N(Up(A)))$
 		\item $ out^{\mathcal{A}}_{II} (N, A) = Up(N(Eq(A)))$
 		\item $ out^{\mathcal{A}}_{1} (N, A) = Up(N(Up(A)))$
 		\item $ out^{\mathcal{A}}_{2} (N, A) =  \bigcap \{ Up(N(V)), A \subseteq V, V \text{is saturated} \}$\footnote{For this case, the abstract logic $\mathcal{A} = \langle \mathbf{Fm}(X), C\rangle $ should include $ \vee $, that is a binary operation symbol, either primitive or defined by a term, and we then have $ a \vee b, b \vee a \in C(\{a\}) $ ($ \vee $-Introduction) and if $ c \in C(\{a\}) \cap C(\{b\}) $ then $ c \in C(a\vee b), C(b\vee a) $ ($ \vee $-Elimination). }
 		%Saturated sets are defined similar as before.
 		\item $ out^{\mathcal{A}}_{3} (N, A) =  \bigcap \{ Up(N(V)), A \subseteq V= Up(V) \supseteq N(V) \}$
 	\end{compactitem}
 	
 	Put $  out^{\mathcal{A}}_{i} (N) =  \{ (A, x) : x \in out^{\mathcal{A}}_{i} (N, A) \} $.
 \end{definition}
 %Given an abstract logic $\mathcal{A} = \langle \mathbf{Fm}(X), C\rangle $ and a normative system $ N \subseteq Fm(X) \times Fm(X)$, it is defined that  $(a,x) \in derive^{\mathcal{A}}_{0}(N) $  ($ derive^{\mathcal{A}}_{I}(N) $, $ derive^{\mathcal{A}}_{II}(N) $, $ derive^{\mathcal{A}}_{1}(N) $,  $ derive^{\mathcal{A}}_{2}(N) $, $ derive^{\mathcal{A}}_{3}(N) $)   if and only if $ (a,x) $ is derivable from $ N $ using the rules $ \{EQI, EQO \} $ \linebreak and ($ \{SI, EQO \} $, $ \{ WO, EQI \} $,   $ \{SI, WO \} $, $ \{SI, WO, OR \} $, $ \{SI, WO, T \} $).
 \begin{definition}[Proof system]
 	Given an abstract logic $\mathcal{A} = \langle \mathbf{Fm}(X), C\rangle $ and a normative system $ N \subseteq Fm(X) \times Fm(X)$, it is defined that  $(a,x) \in derive^{\mathcal{A}}_{i}(N) $     if and only if $ (a,x) $ is derivable from $ N $ using the rules \linebreak$ \{EQI, EQO \} $, $ \{SI, EQO \} $, $ \{ WO, EQI \} $,   $ \{SI, WO \} $, $ \{SI, WO, OR \} $ and $ \{SI, WO, T \} $ for $ i\in \{0, I, II, 1, 2, 3\} $ in turn.
 	 Put $ derive^{\mathcal{A}}_{i}(N,A) $  $= \{x : (A,x) \in derive^{\mathcal{A}}_{i}(N)\}$.
 	% $(a,x) \in derive^{\mathcal{L}}_{2}(N) $ if and only if $ (a,x) $ is derivable from $ N $ using the rules $ \{SI, WO, OR \} $, $(a,x) \in derive^{\mathcal{L}}_{3}(N) $ if and only if $ (a,x) $ is derivable from $ N $ using the rules $ \{SI, WO, T \} $.
 \end{definition}
 \begin{theorem} [Soundness and completeness]\label{Ab: th}
 	$ out^{\mathcal{A}}_{i}(N) = derive^{\mathcal{A}}_{i}(N) $.
 \end{theorem}	
 %\begin{proof}
 %	The proofs are the same as soundness and completeness proofs in Theorem \ref{th:out}.
 %\end{proof}
 
 %We can define Simple-I and Simple-II operations over lattices and abstract algebras similarly, for avoiding repetition we skip writing results about these two systems.
 
 A logical system $ \mathbf{L} = \langle L, \vdash_{\mathbf{L}} \rangle$ straightforwardly provides an equivalent abstract logic $  \langle \mathbf{Fm}_{L}, C_{\vdash_{L}}  \rangle $.
 Therefore, an I/O framework can be built over different types of logics including first-order logic, simple type theory, description logic, as well as different kinds of modal logics that are expressive for intentional concepts such as belief and time. 
 
 \begin{example}
 	 In a modal logic system KT, for the conditionals $ N= \{ (p,\Box q), $ $ (q,r), (s, t) \} $ and input set $ A= \{p\} $, we have $ out^{KT}_{3}(N, A)= Up(\Box q, r)  $.
 \end{example}
 
 Moreover, other rules such as AND and CT can be added to the systems in the same way as in Section~\ref{sec: BOO}.  
 %, but there is proof only for simple-minded operation

%%%%%%%%%%%%%%%%%%%%%%%%%%%%%%%%%%%%%%%%%%%%%%%%%
%\subsection{Nested I/O operations} \label{sec: Nes}

\begin{theorem}\label{Nes}
	Every $ out_{i}^{\mathcal{B}}$, and $ out_{i}^{\mathcal{A}} $ operation is a closure operator.
\end{theorem} 
 
Nested I/O operations can be defined. Since $ out_{i}^{\mathcal{B}} $ is a closure operator, it can be defined that $ out_{j}^{\mathcal{A}} (M, out_{i}^{\mathcal{B}} (N) ) $ where $ N \subseteq Ter(B) \times Ter(B) $, $ M\subseteq (Ter(B)\times Ter(B)) \times (Ter(B)\times Ter(B)) $ and in the abstract logic $ \mathcal{A} $ we have $ L= N\times N $ and $ C = out_{i}^{\mathcal{B}} $. The corresponding characterization is $ derive_{j}^{\mathcal{A}} (M, derive_{i}^{\mathcal{B}} (N) ) $. Similarly, nested $ out_{j}^{\mathcal{A}} (M, out_{i}^{\mathcal{A}} (N) ) $ operations can be defined. 
  
%$ out_{j}^{\mathcal{A}} (M, out_{i}^{\mathcal{H}} (N) ) $, $ out_{j}^{\mathcal{A}} (M, out_{i}^{\mathcal{L}} (N) ) $ and $ out_{j}^{\mathcal{A}} (M, out_{i}^{\mathcal{A}} (N) ) $ operations. 
%%%%%%%%%%%%%%%%%%%%%%%%%%%%%%%%%%%%%%%%%%%%%%%%%%
 
\section{Synthesizing normative reasoning and preferences }\label{sec : SNP}
Input/output logic originally developed on top of classical propositional logic~\cite{Makinson00}. This section shows that the extension of propositional logic with a set of conditional norms is sound and complete with respect to the class of Boolean algebras that the corresponding I/O operation holds.
The language of classical propositional logic consists of the connectives $ \mathcal{L}_{C} = \{\wedge, \vee, \neg, \top, \bot \} $.  Let $ X $ be a set of variables; as usual the set of formulas is defined over $ X $ and referred to as $Fm(X)$.\footnote{For the precise definition, the auxiliary symbols brackets $ ),($ are used. Apart from the use of brackets, the formulas over $ X $ are Boolean terms over $ X $: $ Ter(X) $.} The algebra of formulas over $ X $ is a Boolean algebra as follows:
%we can represent the set of classical propositional logic over $ X $ as a Boolean algebra as follows:
\begin{center}
	$ \mathbf{Fm}(X) = \langle Fm(X), \wedge^{\mathbf{Fm}(X)}, \vee^{\mathbf{Fm}(X)}, \neg^{\mathbf{Fm}(X)}, \top^{\mathbf{Fm}(X)}, \bot^{\mathbf{Fm}(X)}   \rangle $
\end{center}

where  $ \wedge^{\mathbf{Fm}(X)} (\varphi, \psi) = (\varphi\wedge\psi) $, $ \vee^{\mathbf{Fm}(X)} (\varphi, \psi) = (\varphi\vee\psi) $, $\neg^{\mathbf{Fm}(X)} (\varphi) =\neg \varphi $, $ \top^{\mathbf{Fm}(X)} = \top$, and $ \bot^{\mathbf{Fm}(X)} =\bot $. It is defined that $ \varphi  \vdash_{C} \psi$ if and only if $ \varphi  \leq \psi$,  and  $\varphi \dashv \vdash_{C} \phi$ if and only if $ \varphi  \leq \psi$ and $ \psi  \leq \varphi$.
%\footnote{$ \vdash_{C} $  is equal to the logical consequence relation in classical propositional logic. }

\begin{definition}
	Let $ N \subseteq Fm(X)  \times Fm(X)  $ where $ X $ is a set of propositional variables. The formula $ (\varphi,\psi) \in derive^{\mathbf{Fm}(X)}_{i}(N) $ applies if and only if $(\varphi,\psi) $ is derivable from $ N $ using $EQO, EQI,$  $ SI, WO, OR , T$ as follows:
%	\newpage
	\begin{multicols}{3}
	\hspace*{-.3in}
			\begin{tabular}{ l c  }
			\hline
			\emph{derive}$^{\mathbf{Fm}(X)}_i$ & Rules   \\
			\hline
			\emph{derive}$^{\mathbf{Fm}(X)}_{R}$  & \{EQO\}   \\	
			\emph{derive}$^{\mathbf{Fm}(X)}_{L}$  & \{EQI\}   \\
			\emph{derive}$^{\mathbf{Fm}(X)}_{0}$  & \{EQI, EQO\}   \\
			\emph{derive}$^{\mathbf{Fm}(X)}_{I}$  & \{SI, EQO\}   \\
			\emph{derive}$^{\mathbf{Fm}(X)}_{II}$  & \{WO, EQI\}   \\
			\emph{derive}$^{\mathbf{Fm}(X)}_{1}$  & \{SI, WO\}   \\
			\emph{derive}$^{\mathbf{Fm}(X)}_{2}$  & \{SI, WO, OR\}   \\
			\emph{derive}$^{\mathbf{Fm}(X)}_{3}$  & \{SI, WO, T\}  \\
			%\emph{deriv}$^{\mathbf{Fm}(X)}_{4}$  &  \{SI, WO, OR, T\}\\
			\hline
		\end{tabular}
		\begin{prooftree}\hspace*{.05in}
			\AxiomC{$(\varphi,\psi)$ } \AxiomC{$\psi \dashv \vdash_{C} \phi$ }
			\LeftLabel{EQO} \BinaryInfC{$(\varphi,\phi)$}
		\end{prooftree}
		\begin{prooftree}\hspace*{.1in}
			\AxiomC{$(\varphi,\psi)$ } \AxiomC{$\varphi \dashv \vdash_{C} \phi$ }
			\LeftLabel{EQI} \BinaryInfC{$(\phi,\psi)$}
		\end{prooftree}
		\begin{prooftree}\hspace*{.2in}
			\AxiomC{$(\varphi,\psi)$ } \AxiomC{$\phi \vdash_{C} \varphi$ }
			\LeftLabel{SI} \BinaryInfC{$(\phi,\psi)$}
		\end{prooftree}
		\begin{prooftree}\hspace*{.2in}
			\AxiomC{$(\varphi,\psi)$ } \AxiomC{$ (\psi, \phi)$}
			\LeftLabel{T} \BinaryInfC{$(\varphi,\phi)$}
		\end{prooftree}
		\begin{prooftree}\hspace*{.2in}
			\AxiomC{$(\varphi,\psi)$ } \AxiomC{$ (\phi, \psi)$}
			\LeftLabel{OR} \BinaryInfC{$(\varphi \vee \phi,\psi)$}
		\end{prooftree}
		\begin{prooftree}\hspace*{.05in}
			\AxiomC{$(\varphi,\psi)$} \AxiomC{$\psi  \vdash_{C} \phi$ }
			\LeftLabel{WO} \BinaryInfC{$(\varphi,\phi)$}
		\end{prooftree}
	
	\end{multicols}
	It is defined that $ (\varGamma,\psi) \in derive^{\mathbf{Fm}(X)}_{i}(N) $ if $ (\varphi,\psi) \in derive^{\mathbf{Fm}(X)}_{i}(N)  $  for some $ \varphi \in \varGamma \subseteq Fm(X)$. Put $ derive^{\mathbf{Fm}(X)}_{i}(N,\varGamma) = \{\psi : (\varGamma,\psi) \in derive^{\mathbf{Fm}(X)}_{i}(N)  \}  $.
\end{definition}

\begin{example}
 For the conditional norm set $N = \{(\top, g), (g, t),  (\neg g, \neg t )  \} $ and input set $ A= \{ \neg g\} $, we have $ out^{\mathbf{Fm}(X)}_{3}(N, A) = Up (t, \neg t, g)$.
\end{example}

Given $ \langle \mathbf{Fm}(X), \vdash_{C}  \rangle $,
let $ \mathcal{B} $ be a Boolean algebra and $ X $ be a set of propositional variables. A valuation on $ \mathcal{B} $ is a function from $ X $ into the universe of $ \mathcal{B} $. Any valuation on $ \mathcal{B} $ can be extended in a unique way to a homomorphism from the algebra $ \mathbf{Fm}(X) $ into $ \mathcal{B} $. A valuation $ V $ on $ \mathcal{B} $ satisfies a formula if $ V(\varphi) = 1_{\mathcal{B}} $, and it satisfies a set of formulas if it satisfies all its elements~\cite{jansana2016algebraic}. 

\begin{definition}
	For any Boolean algebra $ \mathcal{B} $, the consequence relation $ \vDash_{\mathcal{B}} $ can be defined as follows:
	\begin{center}
		$ \varGamma \vDash_{\mathcal{B}} \varphi $ if and only if for any valuation on $ \mathcal{B} $ that  $ V(\varGamma)=1_{\mathcal{B}} $ \linebreak  then $V(\varphi)=1_{\mathcal{B}} $.
	\end{center}
\end{definition}

\begin{definition}
	Let $ \mathbf{BA} $ be the class of all Boolean algebras. The consequence relation $ \vDash_{\mathbf{BA}} $ can be defined as follows:
	\begin{center}
		$ \varGamma \vDash_{\mathbf{BA}} \varphi   $ if and only if for any Boolean algebra $ \mathcal{B} $, $ \varGamma \vDash_{\mathcal{B}} \varphi   $.
	\end{center}
\end{definition}

\begin{theorem}\label{th: Bo}
	For every set of formulas $ \varGamma $ and every formula $ \varphi $,
	\begin{center}
		$ \varGamma \vDash_{\mathbf{BA}} \varphi   $ if and only if $  \varGamma \vdash_{C} \varphi  $.
	\end{center}
\end{theorem}
%\begin{proof}
%	See \cite{Blok,jansana2016algebraic}. 
%\end{proof}

\begin{theorem}\label{th : Fm}
	Let $ X $ be a set of propositional variables and $ N \subseteq \linebreak  Fm(X) \times Fm(X) $. For a given Boolean algebra $ \mathcal{B} $ and a valuation $ V $ on $ \mathcal{B} $, it is defined that  $ N^{V} = \{(V(\varphi), V(\psi)) | (\varphi,\psi) \in $~$N \} $.   We have 
\begin{tcolorbox}	
	\begin{center}
		$ (\varphi,\psi) \in derive^{\mathbf{Fm}(X)}_{i}(N) $  
	\end{center}
	if and only if
	\begin{center}
		$ V(\psi) \in out^{\mathcal{B}}_{i} (N^{V},  \{ V(\varphi)\})$ for every $ \mathcal{B} \in \mathbf{BA} $ and valuation  $V$.
	\end{center}
\end{tcolorbox}
\end{theorem}

%\begin{proof}
%See Appendix B.
%\end{proof}

  The theorem can be extended for arbitrary input set $ \varGamma \subseteq Fm(X) $. Suppose that  $ (\varGamma,\psi) \in derive^{\mathbf{Fm}(X)}_{i}(N) $,  then $ (\varphi,\psi) \in derive^{\mathbf{Fm}(X)}_{i}(N) $ for $ \varphi \in \varGamma $. As above, we have  $ V(\psi) \in out^{\mathcal{B}}_{i} (N^{V},  \{ V(\varphi)\})$ for every $ \mathcal{B} \in \mathbf{BA} $ and valuation  $V$, so that by definition of $ out^{\mathcal{B}}_{i} $, it can be said that  $ V(\psi) \in out^{\mathcal{B}}_{i} (N^{V},  \{V(\varphi) | V(\varphi) \in V(\varGamma) \}   )$ for every $ \mathcal{B} \in \mathbf{BA}$ and valuation $V$.

\begin{theorem}\label{th : AND}
	Let $ X $ be a set of propositional variables and $ N \subseteq \linebreak Fm(X) \times Fm(X) $. For a given Boolean algebra $ \mathcal{B} $ and a valuation $ V $ on $ \mathcal{B} $, it is defined that  $ N^{V} = \{(V(\varphi), V(\psi)) | (\varphi,\psi) \in $~$N \} $.   We have 
	\\
%	\vspace{2cm}
\begin{tcolorbox}	
	\begin{center}
		$ (\varphi,\psi) \in derive^{AND}_{i}(N) $  
	\end{center}
	if and only if
	\begin{center}
		$ V(\psi) \in out^{AND}_{i} (N^{V},  \{ V(\varphi)\})$ for every $ \mathcal{B} \in \mathbf{BA} $ and valuation  $V$.
	\end{center}
\end{tcolorbox}
%\footnote{ $ out^{AND}_{i} (N^{V},  \{ V(\varphi)\})$ is defined for every Boolean algebra $ \mathcal{B} $, see Definition \ref{outAND}.}
\end{theorem}
%\begin{proof}
%See Appendix B.
%\end{proof}

%\pagebreak
\subsection{Consistency check }
Constraints can be added to the derivation systems such that the output set of formulas is consistent with the proposed constraint.
% Adding constraints to input/output logic is required for dealing with contrary-to-duty problems  \cite{Makinson01}. 
 
\begin{definition}
	Let $ X $ be a set of propositional variables and $ N \subseteq \linebreak Fm(X) \times Fm(X) $. Given the constraint $ Con $ that is a set of formulas $ Con \subseteq Fm(X) $, it is defined that $ (\varphi,\psi) \in derive^{Con}_{i}(N) $ if and only if 

	\begin{center}
		$   (\varphi,\psi) \in derive^{\mathbf{Fm}(X)}_{i}(N) \,\, \, \text{and} \, \, \,  Con, \psi \nvdash_{C} \bot $.
	\end{center}
	
	Given a set of $ \varGamma\subseteq Fm(X) $, it is defined that $ (\varGamma,\psi) \in derive^{Con}_{i}(N) $ if $ (\varphi,\psi) \in derive^{Con}_{i}(N) $ for some $ \varphi \in \varGamma $.

\end{definition}

\begin{theorem}\label{th : con}
	Let $ X $ be a set of propositional variables, $ N \subseteq \linebreak Fm(X) \times Fm(X) $, and $ Con \subseteq Fm(X) $. For a given Boolean algebra $ \mathcal{B} $ and a valuation $ V $ on $ \mathcal{B} $, it is defined that  $ N^{V} = $~$\{(V(\varphi), V(\psi)) | (\varphi,\psi) \in N \} $.   We have 
\begin{tcolorbox}	
	\begin{center}
		$ (\varphi,\psi) \in derive^{Con}_{i}(N) $  
	\end{center}
	if and only if
	\begin{center}
		$ V(\psi) \in out^{\mathcal{B}}_{i} (N^{V},  \{ V(\varphi)\})$ for every $ \mathcal{B} \in \mathbf{BA} $ and valuation  $V$ 
	\end{center}
	\quad \quad \quad \quad  and 
	\begin{center}
		for some $ \mathcal{B} \in \mathbf{BA} $, there is a valuation $ V $  such that  $\forall \delta\in Con, \, V(\delta \wedge \psi )= 1_{\mathcal{B}}$.
	\end{center}
	\end{tcolorbox}
\end{theorem}
%\begin{proof}
%	See Appendix B.
%\end{proof}

\subsection{Integrating preferences}
In the normative systems proposed, it is possible to add a preference relation over the set of valuations and define a new conditional theory.
   Conditional obligation sentences are analyzed which have the form $ a > \bigcirc x $, where $ > $ is a (preferential) conditional connective~\cite{hansson1969analysis,Lewis1}. Given the set of obligatory norms $ N^{O} $, and suppose that $ (a,x) \in N^{O} $, the new conditionals are defined as follows:
\begin{center}
	$ a > \bigcirc x $ holds iff $ (a,x) \in derive_{i}(N^{O}) $ and $ a > x $ holds
\end{center}
%\pagebreak

where $derive_{i}(N^{O})$ is an appropriate derivation system for obligation. Intuitively, the modal translation of $ a \rightarrow \bigcirc x $ for $ (a,x) \in derive_{i}(N^{O}) $~\cite{Makinson00,J46} is considered. This makes the definition a compositional definition of monadic obligation operators and conditionals.

For a given set of permissive norms $N^{P}$, by choosing a plausible derivation system for permission--- $derive_{i}(N^{P})$ ---that is similar to the definition of a conditional obligation, then conditional permission can be defined as 

\begin{center}
	$ a > P x $ holds iff $ (a,x) \in derive_{i}(N^{P}) $ and $\neg (a > \neg x)  $ holds
\end{center}

where $\neg (a > \neg x)$ is the conditional dual of $ a > x $. The set of new conditional obligations is denoted by $ derive^{O}_{i} $ and the set of new conditional permissions is denoted  by $ derive^{P}_{i} $. This article does not refer to the subscripts or superscripts of the normative system or derivation systems where they are clear from the context or do not affect our discussion.

\begin{definition}
	Let $ X $ be a set of propositional variables and $ MaxC $ the set of all the maximal consistent subsets of $ Fm(X) $. Let $ f\subseteq \linebreak MaxC \times MaxC $ be a relation over elements of $ MaxC $ and \linebreak $ \opf (\varphi) = \{ M \in  MaxC   \,\mid\,  \varphi \in M,\, \forall K \, (\varphi \in K \rightarrow (M,K) \in f ) \}$. It is defined that $ \varphi > \bigcirc \psi \in derive^{O^{H}}_{i}(N) $ if and only if 
	%$ \opf (\varphi) = \{ M \in  MaxC  \,\mid\, \varphi \in M,\, \nexists K (K,M) \in f    \}$
	
	\begin{center}
		$    (\varphi,\psi) \in derive^{\mathbf{Fm}(X)}_{i}(N) \, \, \, \text{and} \, \, \, \forall M \in \opf (\varphi) \, (\psi \in M) $.
	\end{center}
	
	Given a set of $ \varGamma\subseteq Fm(X) $, it is defined that $ \varGamma > \bigcirc \psi \in derive^{O^{H}}_{i}(N) $ if $ \varphi > \bigcirc \psi \in derive^{O^{H}}_{i}(N) $ for some $ \varphi \in \varGamma $.
\end{definition}

\begin{definition}
	Let $ X $ be a set of propositional variables and $ f\subseteq \linebreak MaxC \times MaxC $. A preference Boolean algebra for $ \mathbf{Fm}(X) $ is a structure \linebreak $ M= \langle \mathcal{B}, \mathcal{V},  \succeq_{f}\rangle  $ where:
	\begin{compactitem}
		\item $\mathcal{B}$ is a Boolean algebra,
		\item $ \mathcal{V} = \{V_{i}\}_{i\in I} $ is the set of valuations from $ \mathbf{Fm}(X) $ on $ \mathcal{B} $,
		\item $ \succeq_{f} \subseteq \mathcal{V} \times \mathcal{V} $:  $ \succeq_{f} $ is a betterness or comparative goodness relation over valuations from $ \mathbf{Fm}(X) $ to $ \mathcal{B} $ such that  $ V_{i} \succeq_{f} V_{j} $ iff \linebreak  $ (\{\varphi| V_{i}(\varphi)= 1_{\mathcal{B}}\}, \{\psi| V_{j}(\psi) =1_{\mathcal{B}} \} ) \in f $.
	\end{compactitem}
\end{definition}

No specific properties (like reflexivity or transitivity) are considered for the betterness relation. For a given preference Boolean algebra \linebreak $ M= \langle \mathcal{B}, V,  \succeq_{f}  \rangle $, it is defined that $ opt_{\succeq_{f}} (\varphi) = \{ V_{i} \in \mathcal{V}  \,\mid\, V_{i}(\varphi)= 1_{\mathcal{B}},$ \linebreak $ \forall V_{j} (  V_{j}(\varphi)= 1_{\mathcal{B}} \rightarrow V_{i} \succeq_{f} V_{j}) \}$.

\begin{theorem}\label{th : OH}
	Let $ X $ be a set of propositional variables, where $ N \subseteq \linebreak Fm(X) \times Fm(X) $, and $ f\subseteq MaxC \times MaxC $. For a given Boolean algebra $ \mathcal{B} $ and a valuation $ V $ on $ \mathcal{B} $,  it is defined that  $ N^{V} =$~$ \{(V(\varphi), V(\psi)) | (\varphi,\psi) \in N \} $. We  have
	\begin{tcolorbox}
	\begin{center}
		$ \varphi > \bigcirc \psi \in derive^{O^{H}}_{i}(N) $  
	\end{center}
	if and only if
	\begin{center}
		$ V(\psi) \in out^{\mathcal{B}}_{i} (N^{V},  \{ V(\varphi)\})$ for every $ \mathcal{B} \in \mathbf{BA} $ and valuation  $V$, 
	\end{center}
	\quad \quad \quad \quad  and 
	\begin{center}
		for every preference Boolean algebra $ M= \langle \mathcal{B}, \mathcal{V},  \succeq_{f}  \rangle $, 
		\\ for every valuation $ V_{i} \in opt_{\succeq_{f}} (\varphi) $, \\
		it is the case that  $ V_{i}(\psi)= 1_{\mathcal{B}}$.
	\end{center} 
 \end{tcolorbox}	
\end{theorem}

%\begin{proof}
%	See Appendix B.
%\end{proof}
The theorem can also be rewritten as follows:\footnote{If $ V_{i} \in opt_{\succeq_{f}} (\varphi) $ in $ M= \langle \mathbf{2}, \mathcal{V},  \succeq_{f}  \rangle $, then we have $ V_{i} \in opt_{\succeq_{f}} (\varphi) $ in every preference Boolean algebra $ M= \langle \mathcal{B}, \mathcal{V},  \succeq_{f}  \rangle $.} 
\begin{tcolorbox}
\begin{center}
	$ \varphi > \bigcirc \psi \in derive^{O^{H}}_{i}(N) $  
\end{center}
if and only if
\begin{center}
	$ \psi \in out^{\mathbf{Fm}(X)}_{i} (N,  \{\varphi\})$ and in $ M= \langle \mathbf{2}, \mathcal{V},  \succeq_{f}  \rangle $, 
	\\ for every valuation $ V_{i} \in opt_{\succeq_{f}} (\varphi) $, we have  $ V_{i}(\psi)= 1_{\mathcal{B}}$.
\end{center}
\end{tcolorbox}
\begin{example}\label{Ex:ch}
 For the conditional norm set $N = \{(\top, g), (g, t), (\neg g, \neg t)\}$, the maximal consistent sets can be ordered as follows: the best maximal consistent sets have both $ g $ and $ t $ (type $ s_{1} $); the second best maximal consistence sets are those that have either $ \{g, \neg t\} $ (type $ s_{2} $) or $ \{\neg g, \neg t\} $ (type $ s_{3} $); and the worst maximal consistent sets are those that have $ \{\neg g, t\} $ (type $ s_{4} $). 
 \begin{center}
 	best \quad $ s_{1}  \bullet g, t$
 	\\
 	$---------------$
 	\\ 
 	2nd best \quad $ s_{2}  \bullet g \quad s_{3}  \bullet $
 	\\
 	$---------------$
 	\\
 	worst  \quad $ s_{4}  \bullet t $ 
 	\\
 \end{center}
 
 Since $ \forall M \in \opf (\top) \, ( g \in M) $, $ \forall M \in \opf (g) \, ( t \in M) $ and  $ \forall M \in \opf (\neg g) \, ( \neg t \in M) $, we have $\top > \bigcirc g, g > \bigcirc t, \neg g > \bigcirc \neg t  \in derive^{O^{H}}_{i}(N)$.
\end{example}
%%%%%%%%%%%%%%%%%%%%%%%%%%%%%%%%%%%%%%%%%%%%%%
 
\subsection{Integrating preferences along premise sets}

A preference relation can be introduced over  a set of valuations by means of a  premise set~\cite{lewis1981ordering, kratzer2012modals}. Valuations play the role of possible worlds here. 
 
\begin{definition}
Let $ X $ be a set of propositional variables and $ MaxC $ the set of all maximal consistent subsets of $ Fm(X) $. For $ A \subseteq Fm(X) $, let $ f^{A} \subseteq  MaxC \times MaxC $ such that $ f^{A} = \{(K,M)  | \forall \varphi \in A, (\varphi \in M \rightarrow \varphi \in K)  \} $ is a relation over elements of $ MaxC $. Let $ opt_{f^{A}} (\varphi) = \{ M \in  MaxC   \,\mid\, $\linebreak $  \varphi \in M,\, \forall K \, (\varphi \in K \rightarrow (M,K) \in f^{A} ) \}$. It is defined that $ \varphi > \bigcirc \psi \in derive^{O^{K}}_{i}(N) $  if and only if

	\begin{center}
		$  (\varphi,\psi) \in derive^{\mathbf{Fm}(X)}_{i}(N) \, \, \, \text{and} \, \, \, \forall M \in opt_{f^{A}} (\varphi) \, (\psi \in M) $.
	\end{center}

	Given a set of $ \varGamma\subseteq Fm(X) $, it is defined that $ \varGamma > \bigcirc \psi \in derive^{O^{K}}_{i}(N) $ if $ \varphi > \bigcirc \psi \in derive^{O^{K}}_{i}(N) $ for some $ \varphi \in \varGamma $.
\end{definition}

\begin{definition}
	Let $ X $ be a set of propositional variables and $A \subseteq Fm(X)$. A factual-preference Boolean algebra for $\mathbf{Fm}(X)$ is a structure $ M= \langle \mathcal{B}, \mathcal{V},  \succeq_{A}\rangle $, where:
	\begin{compactitem}
		\item $\mathcal{B}  $ is a Boolean algebra,
		\item $ \mathcal{V} = \{V_{i}\}_{i\in I} $ is the set of valuations from $ \mathbf{Fm}(X) $ on $ \mathcal{B} $,
		\item $ \succeq_{A} \, \subseteq \mathcal{V} \times \mathcal{V} $ such that ($ V_{i} \succeq_{A} V_{j} $  iff  $ \forall \varphi \in A \quad ( V_{j}(\varphi) = 1_{\mathcal{B}} \rightarrow \linebreak V_{i}(\varphi) = 1_{\mathcal{B}} $)).
	\end{compactitem}
\end{definition}

Here, the betterness relation is reflexive and transitive by definition.
For a given preference Boolean algebra $ M= \langle \mathcal{B}, V,  \succeq_{A}  \rangle $, it is defined that \linebreak $ opt_{\succeq_{A}} (\varphi) = \{ V_{i} \in \mathcal{V}  \,\mid\, V_{i}(\varphi)= 1_{\mathcal{B}}, \forall \, V_{j} (  V_{j}(\varphi)= 1_{\mathcal{B}} \rightarrow V_{i} \succeq_{A} V_{j}) \}$. 

%$ opt_{\succeq_{f}} (\varphi) = \{ V_{i} \in \mathcal{V}  \,\mid\, V_{i}(\varphi)= 1_{\mathcal{B}}, \forall V_{j} (  V_{j}(\varphi)= 1_{\mathcal{B}} \rightarrow V_{i} \succeq_{f} V_{j}) \}$
\begin{theorem}\label{th : OK}
	Let $ X $ be a set of propositional variables, where $ N \subseteq \linebreak  Fm(X) \times Fm(X) $, and $ A \subseteq Fm(X) $. For a given Boolean algebra $ \mathcal{B} $ and a valuation $ V $ on $ \mathcal{B} $, it is defined that  $ N^{V} = \{(V(\varphi), V(\psi) | (\varphi,\psi) \in N \} $. We  have
		\begin{tcolorbox}
	\begin{center}
		$ \varphi > \bigcirc \psi \in derive^{O^{K}}_{i}(N) $  
	\end{center}
	if and only if
	\begin{center}
		$ V(\psi) \in out^{\mathcal{B}}_{i} (N^{V},  \{ V(\varphi)\})$ for every $ \mathcal{B} \in \mathbf{BA} $ and valuation  $V$  
	\end{center}
	\quad \quad \quad \quad  and 
	\begin{center}
		for every factual-preference Boolean algebra $ M= \langle \mathcal{B}, \mathcal{V},  \succeq_{A}  \rangle $, 
		\\ for every valuation $ V_{i} \in opt_{\succeq_{A}} (\varphi) $, \\
		it is the case that $ V_{i}(\psi)= 1_{\mathcal{B}}$.
	\end{center} 
	\end{tcolorbox}
\end{theorem}

The theorem can be rewritten as follows:\footnote{By the definition of $opt_{f^{A}} (\varphi)$, then $ \psi $ is true in all maximal consistent subsets that include both $ A $ and $ \varphi $, or when $ \varphi $ is inconsistent with $ A $, in all maximal consistent subsets that include $ \varphi $.}
\begin{tcolorbox}
\begin{center}
	$ \varphi > \bigcirc \psi \in derive^{O^{K}}_{i}(N) $  
\end{center}
if and only if
\begin{center}
	$ \psi \in out^{\mathbf{Fm}(X)}_{i} (N,  \{\varphi\})$, and for $ M= \langle \mathbf{2}, \mathcal{V},  \succeq_{A}  \rangle $, 
	\\ for every valuation $ V_{i} \in opt_{\succeq_{A}} (\varphi) $, we have $ V_{i}(\psi)= 1_{\mathcal{B}}$ 
\end{center}
\quad \quad or

\begin{center}
	$ \psi \in out^{\mathbf{Fm}(X)}_{i} (N,  \{ \varphi\})$, and if $ \varphi $ is consistent with $ A $\\ then $ A, \varphi \vdash \psi $, and if $ \varphi $ is inconsistent with $ A $, then $\varphi \vdash \psi $.
\end{center}
\end{tcolorbox}

The results for the constrained assumptions and preferences can be extended for the other systems introduced, for instance $ derive^{AND}_{i}(N) $. 
%Chapter \ref{Chapter3}section 4in
\begin{example}
 For the conditional norm set $N = \{(\top, g), (g, t), (\neg g, \neg t )\}$ and premise set $A= \{ \neg g, \neg g \rightarrow \neg t\}$, the best maximal consistent set is type $ s_{3} $ (see Example \ref{Ex:ch}), while types $ s_{1}, s_{2} $ and $ s_{4} $ are second best. 
 
 \begin{center}
 	best \quad  $s_{3} \bullet $
 	\\
 	$---------------$
 	\\ 
 	2nd best \quad $ s_{2}  \bullet g \quad    $ $ s_{1}  \bullet g, t \quad $ $ s_{4}  \bullet t $
 	\\
 	
 \end{center} 
 Since $ \forall M \in opt_{f^{A}} (\neg g) \, (\neg t \in M) $, we have $ \neg g > \bigcirc \neg t  \in derive^{O^{K}}_{i}(N)$.
 %(Also $ A, \{ \neg g\} \vdash $~$\neg t $)
\end{example}

It is straightforward to rewrite the theorems for conditional permissions.
\begin{multicols}{2}
%	 \begin{minipage}{.4\textwidth}
		\begin{center}
			$ \varphi > P \psi \in derive^{P^{K}}_{i}(N) $  
			\\
			if and only if
			\\
			$ (\varphi,\psi) \in derive^{\mathbf{Fm}(X)}_{i}(N) $ and 
			\\
			for every factual-preference Boolean algebra $ M= \langle \mathcal{B}, \mathcal{V},  \succeq_{A}  \rangle $, 
			\\ there is a valuation $ V_{i} \in opt_{\succeq_{A}} (\varphi) $ such that $ V_{i}(\psi)= 1_{\mathcal{B}}$.
		\end{center}
%   \end{minipage}
%   \begin{minipage}{.4\textwidth}
   		\begin{center}
   		$ \varphi > P \psi \in derive^{P^{H}}_{i}(N) $  
   		\\
   		if and only if
   		\\
   		$ (\varphi,\psi) \in derive^{\mathbf{Fm}(X)}_{i}(N) $ and
   		\\
   		for every preference Boolean algebra $ M= \langle \mathcal{B}, \mathcal{V},  \succeq_{f}  \rangle $, 
   		\\ there is a valuation $ V_{i} \in opt_{\succeq_{f}} (\varphi) $ such that  $ V_{i}(\psi)= 1_{\mathcal{B}}$.
   	\end{center}
%\end{minipage}

\end{multicols}
%%%%%%%%%%%%%%%%%%%%%%%%%%%%%%%%%%%%%%%%%%%
\section{Semantical embedding of input/output logic into HOL } \label{sec: BIO}
The simple type theory developed by Church~\cite{Church40}, a.k.a. classical higher-order logic (HOL), is an expressive language for representing mathematical structures. The syntax and semantics of HOL are well understood~\cite{J6,J43} (for a brief introduction to HOL see Appendix D). It has roots in Frege's book \emph{``Begriffsschrift, eine der arithmetischen nachgebildete Formelsprache des reinen Denkens''}~\cite{Frege1879}  and  Russell's ramified theory of types~\cite{Russell08}. \linebreak The so-called \textit{shallow semantical embedding} approach was developed by \linebreak Benzm\"{u}ller~\cite{J41} for translating (the semantics of) classical and non-classical logics into HOL. Examples include propositional and quantified multimodal logics~\cite{J23,C37} and dyadic deontic logics~\cite{C71,J45}.

Benzm{\"u}ller et al.~\cite{J46} devised an indirect approach to embedding two I/O operations in modal logic and consequently into HOL. One advantage of building I/O operations over Boolean algebras is that the I/O logic can be directly embedded in HOL. For normative system $N$, the structure \linebreak $ \mathcal{N} = \langle \mathcal{B} , V, N^V \rangle$ is called a Boolean normative model, where  $V$ is a valuation from $\mathbf{Fm}(X)$ to $\mathcal{B}$. The semantical embedding of I/O logic is based on Theorem~\ref{th : Fm}, which states that $ (\varphi,\psi) \in derive^{\mathbf{Fm}(X)}_{i}(N) $ holds if and only if $ V(\psi) \in out^{\mathcal{B}}_{i} (N^{V},  \{ V(\varphi)\})$  holds in all Boolean normative models. 

The remainder of this section shows how the embedding  works, abbreviating type $i \typearrow o$ as $\proptype$. The HOL signature is assumed to contain the constant symbols $ N_{\itype \typearrow \proptype}$, $  \neg_{i \typearrow i}$, $ \vee_{i \typearrow i\typearrow i}  $, $ \wedge_{i \typearrow i\typearrow i} $, $ \top_{i} $ and $ \bot_{i} $. Moreover, for each atomic propositional symbol $p^j \in X$ of $\mathbf{Fm}(X)$, the HOL signature must contain a respective constant symbol $p_i^j$. Without loss of generality, it is assumed that besides
those symbols and the primitive logical connectives of HOL, no other constant symbols are given in the signature of HOL.

The mapping $\lfloor \cdot \rfloor$ translates element
$\varphi \in \mathbf{Fm}(X)$ into HOL terms $\lfloor \varphi \rfloor$ of type
$\itype$. The mapping is recursively defined:
\[
\begin{array}{lcl}
\lfloor p^j \rfloor &=& p_i^j \quad \quad p^j \in X\\
%\lfloor A \rfloor &=& A_{\proptype} \quad \quad A \subseteq L\\
\lfloor \top \rfloor &=& \top_i  \\
\lfloor \bot \rfloor &=& \bot_i \\
\lfloor \neg  \varphi \rfloor &=&   \,  \neg_{i \typearrow  i} (\lfloor \varphi \rfloor)  \\
\lfloor  \varphi \vee \psi  \rfloor &=&   \,   \vee_{i \typearrow i\typearrow i}  \lfloor \varphi \rfloor \lfloor \psi \rfloor
\\
\lfloor  \varphi \wedge \psi \rfloor &=&   \,    \wedge_{i \typearrow i\typearrow i} \lfloor \varphi \rfloor \lfloor \psi \rfloor
\\
\lfloor d_{i}(N) (\varphi,\psi)\text{\footnotemark}  \rfloor &=&   (\bigcirc_{i}(N )_{\proptype \typearrow \proptype} \{\lfloor \varphi \rfloor  \} ) \lfloor \psi \rfloor \\
\end{array}
\]
$\bigcirc_{I}(N)_{\proptype \typearrow \proptype} $,
$\bigcirc_{II}(N)_{\proptype \typearrow \proptype} $,
$\bigcirc_{1}(N)_{\proptype \typearrow \proptype} $, $\bigcirc_{2}(N)_{\proptype \typearrow \proptype} $ and  $ \bigcirc_{3}(N)_{\proptype \typearrow \proptype} $
are thereby abbreviated HOL terms:  
%, $\bigcirc_{4}(N)_{\proptype \typearrow \proptype} $  
\[
\begin{array}{ll}
\bigcirc_{I}(N)_{\proptype \typearrow \proptype} &= \lambda A_\proptype \lambda X_\itype (\exists U \,  ( \exists Y \, (\exists Z \, (A \, Z \wedge Z = Y \wedge N \, Y \, U \wedge U \leq X )    )) )
\\
\bigcirc_{II}(N)_{\proptype \typearrow \proptype} &= \lambda A_\proptype \lambda X_\itype (\exists U \,  ( \exists Y \, (\exists Z \, (A \, Z \wedge Z \leq Y \wedge N \, Y \, U \wedge U = X )    )) )
\\
\bigcirc_{1}(N)_{\proptype \typearrow \proptype} &= \lambda A_\proptype \lambda X_\itype (\exists U \,  ( \exists Y \, (\exists Z \, (A \, Z \wedge Z \leq Y \wedge N \, Y \, U \wedge U \leq X )   )) )
\\
\\
\bigcirc_{2}(N)_{\proptype \typearrow \proptype} &= \lambda A_\proptype \lambda X_\itype (\forall V \, ( Saturated \, V \wedge \forall U ( A \, U \rightarrow V \, U)
\\
& \rightarrow \exists Y \, ( \exists Z \, (Z \leq X \wedge N \, Y \, Z \wedge V \, Y) )  ) )
\\
\\ 
\bigcirc_{3}(N)_{\proptype \typearrow \proptype} &= \lambda A_\proptype \lambda X_\itype (\forall V \, ( \forall U ( A \, U \rightarrow V \, U) \wedge V = Up \, V 
\\
&\wedge \, \forall W (\exists Y (V \, Y \wedge N \,Y \, W) \rightarrow  V \, W  )  
\\
& \rightarrow \exists Y \, ( \exists Z \, (Z \leq X \wedge N \, Y \, Z \wedge V \, Y ) ) ) ) 
\\

\end{array}
\]
where
\[
\begin{array}{ll}
\leq   &= \lambda X_\itype \lambda Y_\itype ( X \wedge_{i \typearrow i\typearrow i}  Y  =   X)
\\
Saturated &= \lambda A_\proptype (\forall X \, \forall Y ( (A \, (X \vee_{i \typearrow i\typearrow i}    Y) \, \rightarrow A \, X  \vee A \, Y ) 
\\
& \wedge    ( A \, X \wedge X \leq Y \rightarrow A \, Y )  ) )
%\\
%& \wedge \forall X ( V \, X \vee V \,  X^{'} ) ) 
\\
Up &= \lambda A_\proptype \lambda X_\itype (\exists Z (A \, Z \wedge Z\leq X) ).
\\
% N \cup (X, Y) &= \lambda U_{i} \lambda Z_{i} ( ( ( (U=X) \wedge(Z=Y)) \rightarrow  \top ) 
% \\
% & \vee ( ( (U\neq X) \vee (Z\neq Y)) \rightarrow  N U Z  ) )
% \\
\end{array}
\]

No further specification is needed for $ N_{\itype \typearrow \proptype}$, $  \neg_{i \typearrow i}$, $ \vee_{i \typearrow i\typearrow i}  $, $ \wedge_{i \typearrow i\typearrow i} $, $ \top_{i} $ and $ \bot_{i} $.
 
\footnotetext{$d_{i}(N) (\varphi,\psi)$ is an abbreviation of $ (\varphi,\psi) \in derive^{\mathbf{Fm}(X)}_{i}(N) $.}
\subsection{Soundness and completeness}
To prove the soundness and completeness, that is, faithfulness, of the above embedding, a
mapping from Boolean normative models into Henkin models is employed.

\begin{definition}[Henkin model ${H}^{\mathcal{N}}$ for Boolean normative model $\mathcal{N}$]\label{def:hm}
	\label{embedding} \sloppy
	For any Boolean normative model $ \mathcal{N} = \langle \mathcal{B}, V, N^{V} \rangle$,
	a corresponding Henkin model $H^\mathcal{N}$ is defined. Thus, let a Boolean normative model  $ \mathcal{N} = \langle \mathcal{B}, V, N^{V} \rangle$ be given. Moreover, assume that the finite set $X= \{p^1,,..., p^m\}$, for $m \geq 1$,  are the only atomic symbols in $\mathbf{Fm}(X)$. The embedding requires the corresponding signature of  HOL to provide constant symbols $p_i^j$  such that $\lfloor p^j \rfloor= p_i^j $.

	 A Henkin model ${H}^{\mathcal{N} } = \langle \{D_\alpha\}_{\alpha \in {T}}, I \rangle$ for
	$\mathcal{N}$ is now defined as follows: $D_\itype$ is chosen as the set of $ B $; all other sets $D_{\alpha\typearrow\beta}$ are
	chosen as (not necessarily full) sets of functions from $D_\alpha$ to
	$D_\beta$.  For all $D_{\alpha\typearrow\beta}$,  the rule that
	every term $ t_{\alpha\typearrow\beta} $ must be denoted in
	$D_{\alpha\typearrow\beta}$ must be obeyed (Denotatpflicht). In particular, it is required that $D_{i}$, $D_{i\typearrow i}$, $D_{i\typearrow i\typearrow i}$ and $D_{ i\typearrow\proptype}$ should contain the elements $ I p_i^j $, $ I \top_{i} $, $ I \bot_{i} $, $I \neg_{i \typearrow i} $, $ I \vee_{i\typearrow i \typearrow i} $, $ I \wedge_{i\typearrow i \typearrow i}$ and $I
	N_{i\typearrow\proptype}$. The interpretation function $I$ of ${H}^{\mathcal{N}}$ is
	defined as follows:
	
	\begin{enumerate}[topsep=1pt,itemsep=0ex,partopsep=1ex,parsep=1ex]
		\item For $j=1,...,m$: $Ip_i^j \in D_{i} $ is chosen such that $Ip_i^j = V(p^j) $ in $\mathcal{N}$.
		\item
		$I \top_{i} \in D_{i} $ is chosen such that  $I \top_{i} = V(\top) $ in $\mathcal{N}$.
		\item
		$I \bot_{i} \in D_{i} $ is chosen such that  $I \bot_{i} = V(\bot) $ in $\mathcal{N}$.
		\item
		$I \neg_{i \typearrow i}   \in D_{i\typearrow i} $ is chosen such that $I (\neg_{i \typearrow i} \, \varphi)  = \psi$ iff $\neg \, V(\varphi)  = V(\psi)$ \linebreak  in $\mathcal{N}$. % and $I p_{\proptype}^{j}(s) = F $ otherwise.
		
		\item
		$I\vee_{i\typearrow i \typearrow i} \in D_{i \typearrow i \typearrow i} $ is chosen such that $I \vee_{i\typearrow i \typearrow i} \varphi \psi = \phi$ iff $ V(\varphi) \vee V(\psi) = V(\phi)$
		in $\mathcal{N}$.
		\item
		$I\wedge_{i\typearrow i \typearrow i} \in D_{i \typearrow i \typearrow i} $ is chosen such that $I \wedge_{i\typearrow i \typearrow i} \varphi \psi = \phi $ iff $ V(\varphi) \wedge V(\psi) = V(\phi) $
		in $\mathcal{N}$.
		\item
		$IN_{i\typearrow\proptype} \in D_{i\typearrow\proptype} $ is chosen 
		such that $I N_{i\typearrow\proptype} \varphi \psi = T $ iff $(V(\varphi),V(\psi)) \in N^V$
		in $\mathcal{N}$. % and  $Iav_{i\typearrow\proptype}(s,u) = F $ otherwise.
		
		\item For the logical connectives $\neg$,$ \wedge $, $\vee$, $\Pi$ and $=$ of
		HOL, the interpretation function $I$ is defined as usual (see Appendix D).
		
	\end{enumerate}
	
	The existence of valuation $V$, which is a Boolean homomorphism from the Boolean algebra $\mathbf{Fm}(X)$ into the Boolean algebra $ \mathcal{B} $, guarantees the existence of $ I $ and its above-mentioned requirements.
	Since it is assumed that there are no other
	symbols (apart from $  \top_{i} $, $ \bot_{i} $, $ \neg_{i \typearrow i} $, $ \vee_{i\typearrow i \typearrow i} $, $ \wedge_{i\typearrow i \typearrow i}$, $ 
	N_{i\typearrow\proptype}$
	as well as $ \neg $, $ \vee $, $ \prod $ and $=$)  in the signature of HOL, 
	$I$ is a total function.
	Moreover, the above construction guarantees
	that ${H}^{\mathcal{N}}$ is a Henkin model: $\langle D,I \rangle$ is a frame,
	and the choice of $I$  in combination with the Denotatpflicht ensures
	that for arbitrary assignments, $g$,  $\|.\|^{{H}^{M},g}$ is a
	total evaluation function.

\end{definition}

\begin{lemma}\label{lemma:1}\sloppy
	Let ${H}^{M} = \langle \{D_\alpha\}_{\alpha \in {T}}, I
	\rangle$ be a Henkin model for Boolean normative model $\mathcal{N}$. We have $
{H}^{\mathcal{N}} \models^\text{HOL} \Sigma $ for all $\Sigma
	\in \{COM \vee, COM \wedge, ASS  \vee,$ 
	\\
	$ASS \wedge, IDE  \vee, IDE  \wedge,$  $ COMP \vee,$  $  COMP \wedge, Dis \vee \wedge, Dis \wedge \vee      \} $, where: \\
	\begin{tabular}{lcl}
		COM$\vee $ & is & $ \forall X_i \, Y_i \, \, ( X \vee Y =  Y \vee X ) $ \\
		COM$\wedge$ & is & $ \forall X_{i} \, Y_i \, \, ( X \wedge Y = Y \wedge X)  $ \\
		ASS$\vee$ & is & $ \forall X_i \, Y_i \, Z_i \, \, ( X \vee (Y \vee Z) =  (X \vee Y) \vee Z ) $ \\
		ASS$\wedge$ & is & $ \forall X_{i} \, Y_i \, Z_i \, \, ( X \wedge (Y \wedge Z) = (X \wedge Y) \wedge Z )  $ \\
		IDE$\vee$ & is & $ \forall X_i \, \, ( X \vee \bot = X ) $ \\
		IDE$\wedge$ & is & $ \forall X_{i} \, \, ( X \wedge \top  = X ) $ \\
		COMP$\vee$ & is & $ \forall X_i \, \, ( X \vee \neg X  =\top ) $ \\
		COMP$\wedge$ & is & $ \forall X_{i} \, \, ( X \wedge \neg X  = \bot ) $ \\
		Dis$\vee\wedge$ & is & $ \forall X_i \, Y_i \, Z_i \, \, ( X \vee (Y \wedge Z) =  (X \vee Y) \wedge (X \vee Z) ) $ \\
		Dis$\wedge\vee $ & is & $ \forall X_{i} \, Y_i \, Z_i \, \, ( X \wedge (Y \vee Z) = (X \wedge Y) \vee (X \wedge Z) )  $ \\
	\end{tabular}
	
\end{lemma}

\begin{lemma}\label{lemmaab:3} 
	Let ${H}^{\mathcal{N}}$ be a Henkin model for Boolean normative model \linebreak $\mathcal{N}= \langle \mathcal{B}, V, N^V \rangle$. For all conditional norms $(\varphi,\psi)$ with arbitrary variable assignments $g$,  it holds that    $ V(\psi)\in out^{\mathcal{B}}_{i} (N, \{V(\varphi)\})  $  if and only if \linebreak $\| \lfloor d_{i}(N) (\varphi,\psi) \rfloor\|^{{H}^{\mathcal{N}}, g }=T$.
\end{lemma}

\begin{lemma} \label{lemmaab4}
	For every Henkin model
	${H} = \langle \{D_\alpha\}_{\alpha \in {T}}, I \rangle$ such that
	${H}\models^\text{HOL} \Sigma $ for all 
	$ \Sigma\in $   $\{\,COM \vee, COM \wedge, ASS  \vee, ASS \wedge, IDE  \vee, IDE  \wedge, COMP \vee, $ \\ $ COMP \wedge, Dis \vee \wedge,$  $ Dis\wedge\vee \,\}$, there exists a
	corresponding Boolean
	normative model  $\mathcal{N}$. Corresponding means that for all conditional norms $ (\varphi,\psi) $ and for all $g$ assignments, then 
	$ \| \lfloor d_{i}(N)(\varphi,\psi) \rfloor \|^{{H}, g } = T
	$ if and only if 
	$V(\psi) \in out^{\mathcal{B}}_{i} (N^V, \{V(\varphi)\}) $.
\end{lemma}

\begin{theorem}[Soundness and completeness of the embedding]\label{th:soco}

	\[
	\begin{array}{ll}
	\text{For every Boolean normative model}   \, \, \, \mathcal{N}, \quad 
	V(\psi) \in out^{\mathcal{B}}_{i} (N^V, \{V(\varphi)\}) 
	\end{array}
	\]
	\text{ if and only if }
	\[
	\begin{array}{ll}
	\{COM \vee,..., Dis \wedge \vee  \} \models^\text{HOL}
	\lfloor d_{i}(N)(\varphi,\psi) \rfloor.
	\end{array}
	\]

\end{theorem}

\pagebreak
\section{Related work}
\label{sec: RW}
 
%%%%%%%%%%%%%%%%%
%In contrast to the earlier input/output logics, we defined non-adjunctive input/output operations. Non-adjunctive logical systems are those where deriving  the conjunctive formula $ \varphi \wedge \psi $ from the set $ \{\varphi, \psi \} $ fails \cite{ciuciura2013non,costa2005non}. 
 %%%%%%%%%%%%%%%%%%%%%%%%%%%%%

 	Gabbay, Parent and van der Torre~\cite{gabbay19} proposed building an I/O framework on top of lattices. They have results only for the simple-minded output operation. This article has shown that for an input set A, by using the  \textit{upward-closed set of A} operator instead of the \textit{upward-closed set of the infimum of A}~\cite{gabbay19}, many new and old derivation systems can be built over Boolean algebras, Heyting algebras, and generally any abstract logic. The algebrization of the I/O framework shows more similarity with the theory of joining-systems~\cite{Lindahl}, an algebraic approach for the study of normative systems over Boolean algebras. It can be said that norms in the I/O framework play the same role as joining in the theory of Lindahl and Odelstal~\cite{Lindahl,Sun}. There are important similarities between input/output logic and the theory of joining-systems, such as studying normative systems as deductive systems and representing norms as ordered pairs. Moreover, both frameworks can generally be built on top of algebraic structures such as Boolean algebras and lattices. While the focus in input/output logic is deontic and factual detachment, the central themes of the theory of joining-systems are intermediate concepts and representing normative systems as a network of subsystems and their inter-relationships (for more details, see \cite{Lindahl}). Sun~\cite{Sun} built Boolean joining systems that characterize I/O logic in a sense that a norm is derivable from a set of norms if and only if it is in the set of norms algebraically generated in the Lindenbaum-Tarski algebra for propositional logic. As in the Bochman approach~\cite{bochman2005explanatory}, the work of Sun~\cite{Sun} has no direct connection to input/output operations. In this article, algebraic I/O operations were built directly over Boolean algebras and, more generally, abstract logics. There is a similar result for building the simple-minded  I/O operation over Tarskian consequence relations in~\cite{carnielli2009input} (see the discussion about abstract input/output logic in  \cite{sun2016logic}).

 This article defined two groups of operations similar to the possible world semantics characterization of box and diamond, where box is closed under AND ($(\Box \varphi \wedge \Box \psi) \rightarrow \Box (\varphi \wedge \psi)$), and not in diamond:
  
 	\hspace{.5cm}	\textit{Derivations systems that do not admit the AND rule}: In the main literature of input/output logic developed by Makinson and van der Torre~\cite{Makinson00},  Parent, Gabbay, and van der Torre~\cite{Parent1}, Parent and van der Torre~\cite{parent2014sing,parent2017pragmatic,parent2018logics}, and Stolpe~\cite{stolpe2008normative,stolpe2008norms,stolpe2015concept}, at least one form of AND inference rule is present. 
 	Sun~\cite{sun2016logic} analyzed norm derivation rules of input/output logic in isolation. Still, it is not clear how to combine them and build new logical systems, specifically systems that do not admit the AND rule. This article has shown how to remove the AND rule from the proof system and build new I/O operations to produce permissible propositions. Unlike minimal deontic logics~\cite{chellas1980modal,Goble13}, and similar approaches such as that of  Ciabattoni, Gulisano and Lellmann~\cite{ciabattoni2018resolving} that do not have deontic aggregation principles, the approach presented in this article validates deontic and factual detachment.

 	\hspace{.5cm}	\textit{Derivations systems that admit the AND rule}: In accordance with the reversibility of inference rules in the I/O proof systems, this article has shown how it is possible to add AND and other rules required for obligation~\cite{Makinson00} to the proof systems, and find I/O operations for them.
       
      There are other abstract approaches: I/O operations over semigroups~\cite{tosatto2012abstract}, which does not admit AND, and a detachment mechanism over an arbitrary set~\cite{ambrossio2017non} that admits a kind of AND, called cumulative aggregation. However, it is not clear how these approaches can be used for logical purposes.

  Constrained I/O logic~\cite{Makinson01} was introduced for reasoning about contrary-to-duty problems. In this article, constraints are preferences. In this sense, the article has presented a semantic characterization for constrained I/O logic. There are 
  syntactic~\cite{Makinson01}, (c.f. Section 6) and proof theoretical~\cite{strasser2016adaptive} (c.f. Section 3) characterizations for constrained I/O logic. The semantic characterization is more flexible than the syntactic characterization mentioned since the approach presented here does not necessarily depend on the AND, SI, and (EQI, EQO) rules required for syntactic characterization of I/O operations in modal logic~\cite{strasser2016adaptive}.

\newpage
 
\section{Conclusion} \label{sec: CN}
 This article presented new algebraic systems developed in the LogiKEy normative reasoning framework. A dataset of semantical embeddings of deontic logics in HOL is available (see Appendix A). The dataset can be used for ethical and legal reasoning tasks. In summary, this article characterizes a class of proof systems over Boolean algebras for a set of explicitly given conditional norms as follows:

\begin{multicols}{3}
\hspace*{-.3in} 
	\begin{tabular}{l c}
		\hline
		\emph{derive}$^{\mathcal{B}}_i$ & Rules   \\
		\hline
		\emph{derive}$^{\mathcal{B}}_{R}$  & \{EQO\}   \\
		\emph{derive}$^{\mathcal{B}}_{L}$  & \{EQI\}   \\	
		\emph{derive}$^{\mathcal{B}}_{0}$  & \{EQI, EQO\}   \\
		\emph{derive}$^{\mathcal{B}}_{I}$  & \{SI, EQO\}   \\
		\emph{derive}$^{\mathcal{B}}_{II}$  & \{WO, EQI\}   \\
		\emph{derive}$^{\mathcal{B}}_{1}$  & \{SI, WO\}   \\
		\emph{derive}$^{\mathcal{B}}_{2}$  & \{SI, WO, OR\}   \\
		\emph{derive}$^{\mathcal{B}}_{3}$  & \{SI, WO, T\}  \\
		%\emph{deriv}$^{\mathcal{B}}_{4}$  &  \{SI, WO, OR, T\}\\
		\hline
	\end{tabular}   
	
   \vspace*{-.2in}
   \begin{prooftree}\hspace*{.1in}
   	\AxiomC{$(a,x)$} \AxiomC{$x = y$ }
   	\LeftLabel{EQO} \BinaryInfC{$(a,y)$}
   \end{prooftree}
   \begin{prooftree}\hspace*{.1in} 
   	\AxiomC{$(a,x)$} \AxiomC{$a = b$ }
   	\LeftLabel{EQI} \BinaryInfC{$(b,x)$}
   \end{prooftree}
   \begin{prooftree}\hspace*{.1in}
   	\AxiomC{$(a,x)$ } \AxiomC{$b\leq a$ }
   	\LeftLabel{SI} \BinaryInfC{$(b,x)$}
   \end{prooftree}

   \begin{prooftree}   
   	\AxiomC{$(a,x)$ } \AxiomC{$ (x, y)$}
   	\LeftLabel{T} \BinaryInfC{$(a,y)$}
   \end{prooftree}
   \begin{prooftree} 
   	\AxiomC{$(a,x)$ } \AxiomC{$ (b, x)$}
   	\LeftLabel{OR} \BinaryInfC{$(a \vee b,x)$}
   \end{prooftree}
   \begin{prooftree} 
   	\AxiomC{$(a,x)$} \AxiomC{$x\leq y$ }
   	\LeftLabel{WO} \BinaryInfC{$(a,y)$}
   \end{prooftree}
\end{multicols}

Each proof system is sound and complete for an input/output (I/O) operation.  The I/O operations resemble inferences, where inputs need not be included among outputs, and outputs need not be reusable as inputs~\cite{Makinson00}.  Moreover, this article has shown how to add the two rules AND and CT to the proof systems and find corresponding I/O operations for them.

\begin{multicols}{2}
	\hspace*{-.3in}
	\begin{tabular}{ll}
		\hline
		\emph{derive}$^{X}_i$ & Rules   \\
		\hline	
		\emph{derive}$^{AND}_{II}$  & \{WO, EQI, AND\}   \\
		\emph{derive}$^{AND}_{1}$  & \{SI, WO, AND\}   \\
		\emph{derive}$^{AND}_{2}$  & \{SI, WO, OR, AND\}   \\
		\emph{derive}$^{CT}_{I}$  & \{SI, EQO, CT\}   \\
		\emph{derive}$^{CT}_{II}$  & \{WO, EQI, CT\}   \\
		\emph{derive}$^{CT}_{1}$  & \{SI, WO, CT\}   \\
		%\emph{deriv}$^{CT,AND}_{I}$  & \{SI, EQO, CT, AND  \} \\
		\emph{derive}$^{CT,AND}_{1}$  & \{SI, WO, CT, AND\}   \\
		\emph{derive}$^{OR}_{I}$  & \{SI, EQO, OR\}   \\
		\emph{derive}$^{CT,OR}_{I}$  & \{SI, EQO, CT, OR\}   \\
		\emph{derive}$^{CT, OR}_{1}$  & \{SI, WO, CT, OR\}   \\
		\emph{derive}$^{CT,OR,AND}_{1}$  & \{SI, WO, CT, OR, AND\}   \\
		\hline
	\end{tabular}
	 
   \begin{prooftree}\hspace*{.2in}
   	\AxiomC{$(a,x)$ } \AxiomC{$ (a,y)$}
   	\LeftLabel{AND} \BinaryInfC{$(a, x\wedge y)$}
   \end{prooftree}
   \begin{prooftree}\hspace*{.2in}
   	\AxiomC{$(a,x)$ } \AxiomC{$ (a\wedge x,y)$}
   	\LeftLabel{CT} \BinaryInfC{$(a,y)$}
   \end{prooftree}

\end{multicols}

The input/output logic is inspired by a view of logic as a \textit{secretarial assistant} tasked with preparing inputs before they go into the motor engine and are unpacked as outputs, rather than logic as an \textit{inference motor}~\cite{Makinson00}.  The only input/output logics investigated so far in the literature are built on top of classical propositional logic and intuitionist propositional logic~\cite{Parent1}. The algebraic construction presented has shown how to build input/output operations on top of any abstract logic.

Finally, this article has proved that the extension of propositional logic with a set of conditional norms is sound and complete with respect to the class of Boolean algebras, and that the corresponding I/O operation holds for all of them. Based on this result, a conditional theory has been integrated into input/output logic.

\section*{Acknowledgments}
  
I would like to thank Leon van der Torre, Dov Gabbay, Christoph Benzm\"{u}ller, Xavier Parent, and Majid Alizadeh for comments that greatly improved the manuscript. Thanks also to Llio Humphreys for her proofreading of the English in this article.

  \bibliographystyle{elsarticle-num} 
  \bibliography{Bibliography.bib}
  \pagebreak
  \section*{Appendix A} \label{sec: IMB}
The semantical embedding outlined in Section \ref{sec: BIO} has been
implemented in the higher-order proof assistant Isabelle/HOL~\cite{Isabelle}. Figures 1 and 2 display their respective encoding.
 Figure 1, after introducing type $ i $ for representing the elements of Boolean algebra, introduces the  algebraic operators as constants in higher-order logic. The algebraic operators are also characterized in accordance with the definition of Boolean algebra.
 
\newtcolorbox[blend into=figures]{myfigure}[2][left=0pt,
      right=0pt,
      top=0pt,
      bottom=0pt,
      width=\textwidth,
      enlarge left by=-2mm,
      boxsep=4pt,
      arc=3pt,
      outer arc=3pt]{float=h!,capture=hbox,title={#2}, every float=\centering, #1,label=#1}
 
\begin{myfigure}{Semantical embedding of Boolean algebra in Isabelle/HOL}\label{fig:bool}
\includegraphics[width=\linewidth]{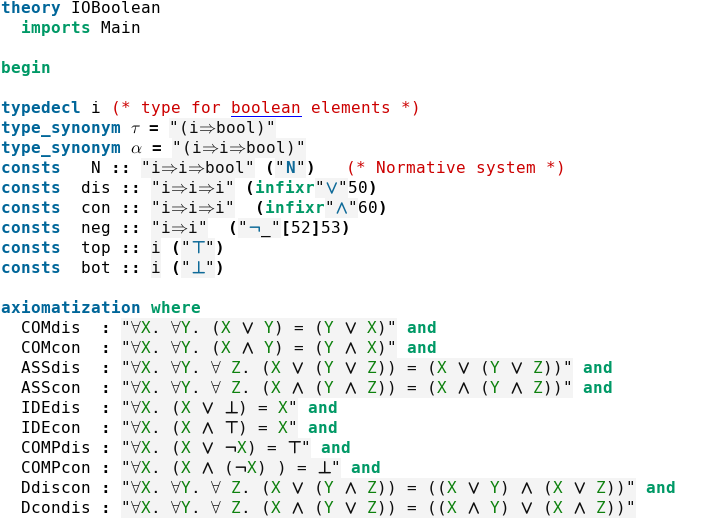}
\end{myfigure}

Figure 2 displays the semantical embedding of I/O operations ($out_i$) in HOL, including the definition of the upward-closed set operator and  saturated set.

% \vspace{-1cm}
\begin{myfigure}{Semantical embedding of $out_i$ in Isabelle/HOL}\label{fig:out}
\includegraphics[width=\linewidth]{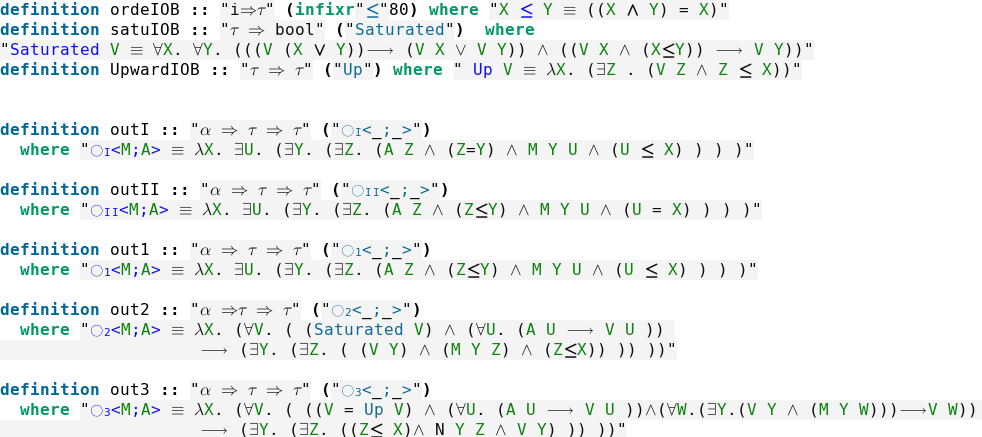}
\end{myfigure}

%\pagebreak 
Figure 3 shows some experiments via the model and countermodel finder Nitpick~\cite{Nitpick}, and prove some facts about I/O operations using automatic theorem provers (\textit{auto} and \textit{meson}) via the Sledgehammer tool~\cite{blanchette2013extending}.

\begin{myfigure}{Some experiments on  $out_i$ in Isabelle/HOL}\label{fig:lemma}
\includegraphics[width=\linewidth]{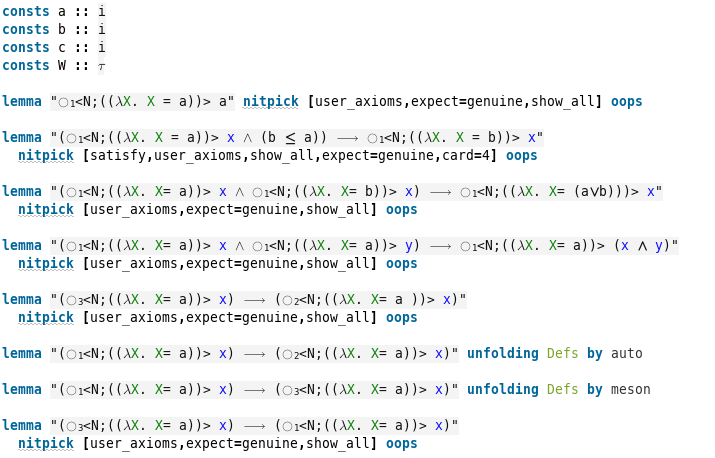}  
\end{myfigure}

%\linebreak
In Figure 4, the first two lemmas prove the soundness of $ out_{1} $. The next two lemmas show the factual detachment of this output operation. The last two lemmas illustrate the soundness of $out_I$ and $out_{II}$ where the depth of inference is one.

  \vspace*{\floatsep}% https://tex.stackexchange.com/q/26521/5764

\begin{myfigure}{Soundness of $out_1$ in Isabelle/HOL}\label{fig:out1}
\includegraphics[width=\linewidth]{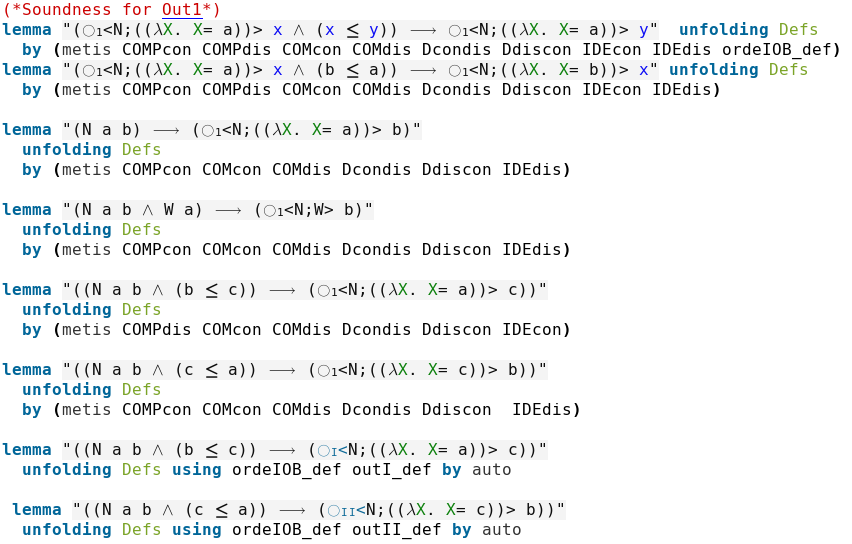}
\end{myfigure}
 Figure 5 shows the soundness of $ out_{2} $ and $ out_{3} $ for a depth of one. The input/output operations introduced by Makinson and van der Torre~\cite{Makinson00} are implemented in Figure 6. The implementations are based on the \textit{reversibility of rules in the derivation systems}. The four input/output operations introduced in~\cite{Makinson00} were built over the \textit{simple-minded output operation} (see Section~\ref{sec: BOO}). 
  
 \pagebreak
 \begin{compactitem}
 \item[]
\begin{myfigure}{Soundness of $out_2$ and $out_3$ in Isabelle/HOL}\label{fig:out23} 
	\includegraphics[width=\linewidth]{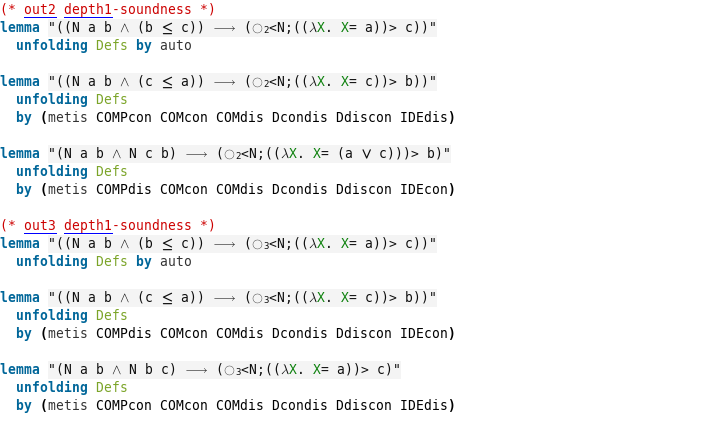}
\end{myfigure}
\item[]
\begin{myfigure}{Semantical embedding of output operations in Isabelle/HOL}\label{fig:outorg}
	\includegraphics[width=\linewidth]{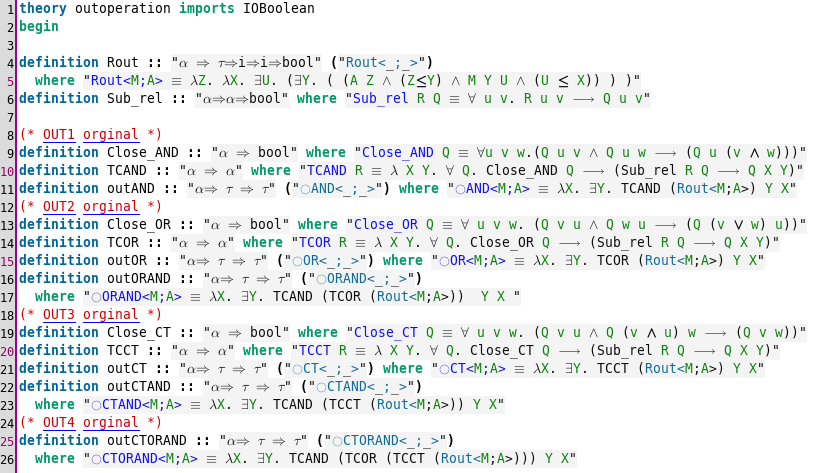}
\end{myfigure}
\end{compactitem}
\newpage

%\pagebreak

The following lemmas (see ~Fig.~7) show the automation capability of implemented output operations for the \textit{simple-minded output operation} \linebreak  ($ out^{AND}_{1} (N) $) as introduced by Makinson and van der Torre~\cite{Makinson00}. 
 
 \begin{myfigure}{Semantical embedding of output operations in Isabelle/HOL}\label{fig:outorgex}
	\includegraphics[width=\linewidth]{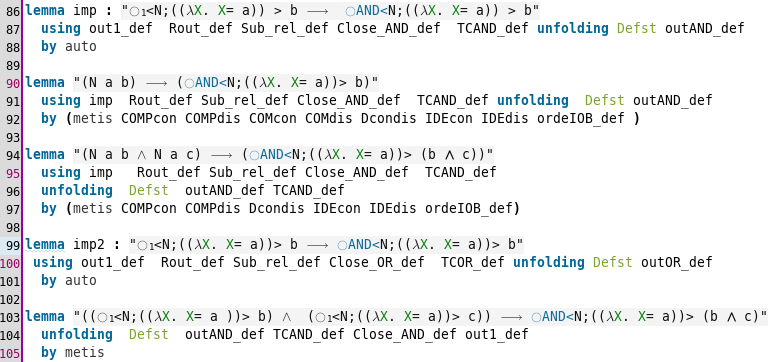}
\end{myfigure}

 The proof system of input/output logic can be implemented directly in Isabelle/HOL---see ~Fig.~8 and 9. The idea is based on an (universal) order of rules in a derivation. The ordering of rules and closure operation are the main ways of defining the derivation systems (for more details, see Section~\ref{sec: BOO}.) For example, in line 27 of Fig.9, \texttt{derSIEQO} introduces the derivation system $derive_I$ with the rules in $\{SI, EQO\}$ and in lines 51--52, \texttt{derSIWOCTORAND} introduce the derivation system $derive_4$ with the rules in $\{SI,WO, CT, OR, AND\}$.
 
 \begin{myfigure}{Semantical embedding of I/O proof systems in Isabelle/HOL}\label{fig:outpro1}
	\includegraphics[width=\linewidth]{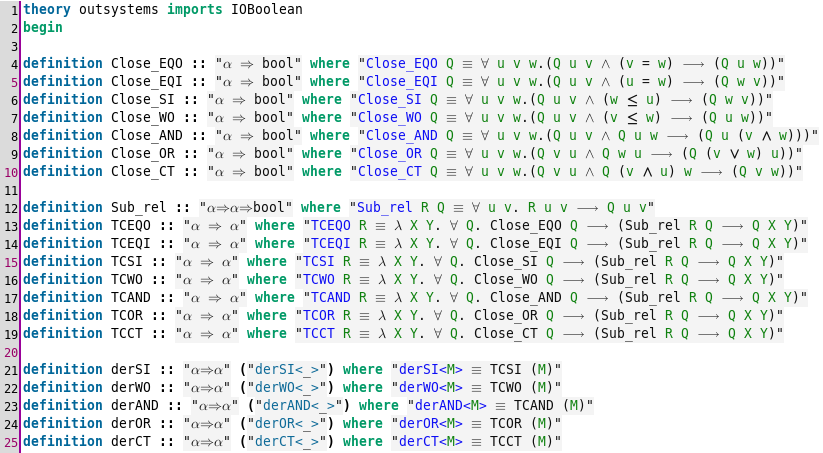}
\end{myfigure}
 
\begin{myfigure}{Semantical embedding of I/O proof systems in Isabelle/HOL}\label{fig:outpro2}
	\includegraphics[width=\linewidth]{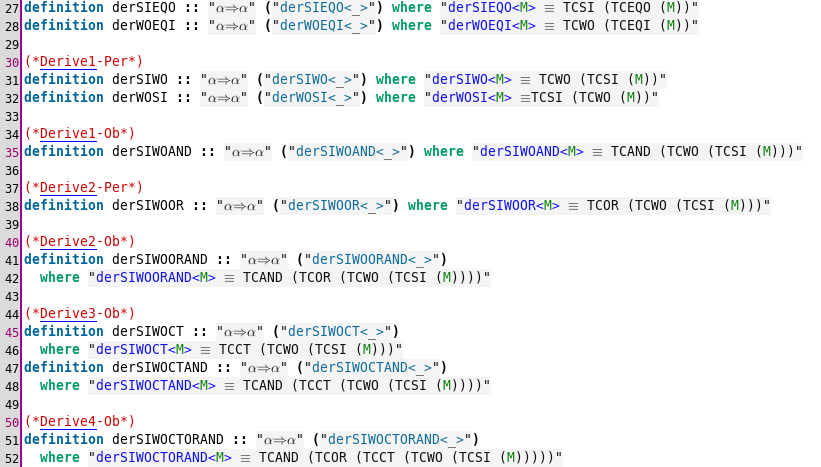}
\end{myfigure}

\pagebreak

One advantage of implementing the proof system of I/O logic, besides the output operations, is that completeness theorems can be checked. For example, the completeness of $out1$, as shown in Fig.~10, is checked in lines 70--73. Lines 61 and 62 show the AND closure. Lines 64--67 demonstrate automation of the implementation for a normative system $M$.
 
 %Lines 64--67 prove its capability of implementation for a normative system $M$.
\begin{myfigure}{Completeness checking of $out1$ in Isabelle/HOL}\label{fig:outprocom}
	\includegraphics[width=\linewidth]{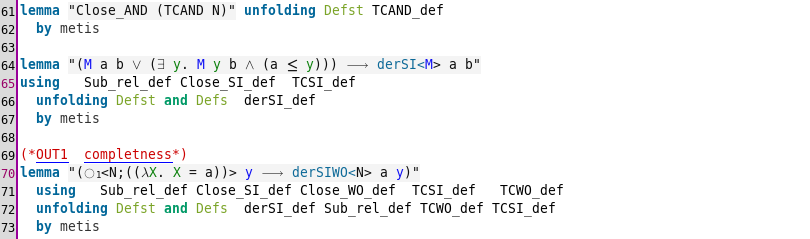}
\end{myfigure}

The proof theoretical difference of different I/O systems can be examined (cf.~Fig.~11). For example, lines 81--85 show that the implemented derivation system \texttt{derSIWOOR} ($derive_2$) is sound for the OR rule for a depth of one.

\begin{myfigure}{Some experiments on I/O proof systems in Isabelle/HOL}\label{fig:outproex}
	\includegraphics[width=\linewidth]{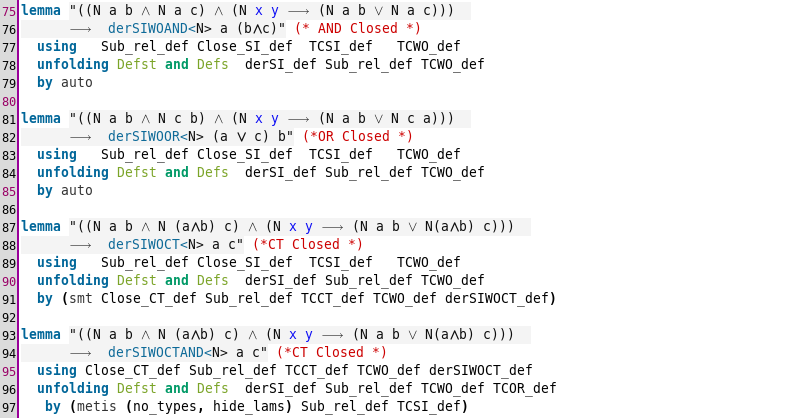}
\end{myfigure}

 \pagebreak 
 \section*{Appendix B} \label{sec:A}  
 
  \subsection*{Proof for Theorem \ref{th:out}: Zero Boolean I/O operation }
  
  Outline of proof for soundness: for the input set $ A \subseteq Ter(B) $, it is shown that if $(A,x) \in derive^{\mathcal{B}}_{0}(N)$, then $ x \in out^{\mathcal{B}}_{0}(N, A)$. By definition, $(A,x) \in derive^{\mathcal{B}}_{0}(N)$  iff $(a,x) \in derive^{\mathcal{B}}_{0}(N)$ for some $ a \in A $. By induction on the length of derivation and since $ out^{\mathcal{B}}_{0} (N) $ validates $EQI$ and $EQO$, if $ (a,x) \in derive^{\mathcal{B}}_{0}(N) $, then $ x \in out^{\mathcal{B}}_{0}(N, \{a\}) $. Thus, by definition of $out^{\mathcal{B}}_{0}$, we have  $ x \in out^{\mathcal{B}}_{0}(N, A)$. If $A=\{\}$, then by definition $ (A,x) \notin derive^{\mathcal{B}}_{0}(N)$. 
 The outline works for the soundness of other systems presented in this appendix as well.
 
 \subsubsection*{Soundness: $ out^{\mathcal{B}}_{0} (N) $ validates $EQI$ and $EQO$.}
 
 	\begin{itemize}
 		\item[EQI:] It needs to be shown that 
 		\begin{prooftree}
 			\AxiomC{$x \in  Eq(N(Eq(a))) $ } \AxiomC{$a=b$ }
 			\LeftLabel{EQI} \BinaryInfC{$x \in Eq(N(Eq(b)))$}
 		\end{prooftree}
 		If $ x \in  Eq(N(Eq(a)))$, then there are $ t_{1} $ and $ t_{2} $ such that $ t_{1} = a $ and $ t_{2} = x $ and $ (t_{1}, t_{2}) \in N$. If $a=b$ then $ t_{1} = b $. Hence, by definition, $ x \in  Eq(N(Eq(b)))$.	 
 		\item[EQO:] It needs to be shown that 
 		\begin{prooftree}
 			\AxiomC{$x \in  Eq(N(Eq(a))) $ } \AxiomC{$x=y$ }
 			\LeftLabel{EQO} \BinaryInfC{$y \in  Eq(N(Eq(a)))$}
 		\end{prooftree}
 		If $ x \in  Eq(N(Eq(a)))$, then there are $ t_{1} $ and $ t_{2} $ such that $ t_{1} = a $ and $ t_{2} = x $ and $ (t_{1}, t_{2}) \in N$. If $x=y$ then $ t_{2} = y $. Hence, by definition, $ y \in  Eq(N(Eq(a)))$.	 
 		
 	\end{itemize}	
 
  \subsubsection*{Completeness: $  out^{\mathcal{B}}_{0}(N) \subseteq derive^{\mathcal{B}}_{0}(N) $.\footnote{For the completeness proofs,  if $ A= \{\} $, then by definition of $ Eq(\{\}) =\{\} $ and $ Up(\{\}) =\{\} $, we have $ x \notin out^{\mathcal{B}}_{i}(N, \{\}) = \{\} $.} }
 	It is shown that if $x \in  Eq(N(Eq(A)))$, then $ (A,x)  \in derive^{\mathcal{B}}_{0}(N)$. Suppose that $x \in  Eq(N(Eq(A)))$, then there are $ t_{1} $ and $ t_{2} $ such that $ t_{1} = a $ and $ a \in A $, and $ t_{2} = x $ such that $ (t_{1}, t_{2}) \in N$.
 	
 	\begin{prooftree} 
 		\AxiomC{$  (t_{1}, t_{2})  $}
 		\AxiomC{$  t_{2} = x $ }
 		\LeftLabel{$EQO$}
 		\BinaryInfC{$  (t_{1}, x)  $}
 		\AxiomC{$ t_{1} = a$ }
 		\LeftLabel{$EQI$}
 		\BinaryInfC{$ (a,  x) $}
 	\end{prooftree}
 	
 	Thus, $ x \in derive^{\mathcal{B}}_{0}(N, a)$ and then $ x \in derive^{\mathcal{B}}_{0}(N, A) $.
  \subsection*{Proof for Theorem \ref{th:out}: Simple-I Boolean I/O operation }	
  \subsubsection*{Soundness: $ out^{\mathcal{B}}_{I} (N) $ validates $ SI $ and $EQO$.}
  \begin{itemize}
 		\item[SI:] It needs to be shown that
 		\begin{prooftree}
 			\AxiomC{$x \in  Eq(N(Up(a))) $ } \AxiomC{$b\leq a$ }
 			\LeftLabel{SI} \BinaryInfC{$x \in Eq(N(Up(b)))$}
 		\end{prooftree}
 		If $ x \in Eq(N(Up(a)))$, then $ \exists t_{1} $ such that $ a \leq t_{1} $ and $ (t_{1}, x) \in N$ or ($ (t_{1}, y) \in N$ and $ y = x $). Hence, if $ b \leq a $, we have $ b \leq t_{1} $ and then $x \in Eq(N(Up(b)))$.	 
 		\item[EQO:] It needs to be shown that 
 		\begin{prooftree}
 			\AxiomC{$x \in  Eq(N(Up(a))) $ } \AxiomC{$x=y$ }
 			\LeftLabel{EQO} \BinaryInfC{$y \in  Eq(N(Up(a)))$}
 		\end{prooftree}
 		If $ x \in  Eq(N(Up(a))) $, then by definition of $  Eq(X) $, if $ x=y $, we have $ y \in  Eq(N(Up(a))) $.
 		
 	\end{itemize}	
  \subsubsection*{Completeness: $  out^{\mathcal{B}}_{I}(N) \subseteq derive^{\mathcal{B}}_{I}(N) $.}
 		It is shown that if $x \in  Eq(N(Up(A)))$, then $ (A,x)  \in derive^{\mathcal{B}}_{I}(N)$. Suppose that $x \in  Eq(N(Up(A)))$, then there is $ t_{1} $ such that $ a \leq t_{1} $ and  $ (t_{1}, x) \in N$  or ($ (t_{1}, y) \in N$ and $ y = x $)  for $ a \in A$. There are two cases: 
 	
 	\begin{multicols}{2}
 		\begin{prooftree} 
 			\AxiomC{$  (t_{1}, x)  $}
 			\AxiomC{$ a \leq t_{1} $ }
 			\LeftLabel{$SI$}
 			\BinaryInfC{$ (a,  x) $}
 		\end{prooftree}
 		\begin{prooftree} 
 			\AxiomC{$  (t_{1}, y)  $}
 			\AxiomC{$ y = x $ }
 			\LeftLabel{$EQO$}
 			\BinaryInfC{$  (t_{1}, x)  $}
 			\AxiomC{$ a \leq t_{1} $ }
 			\LeftLabel{$SI$}
 			\BinaryInfC{$ (a,  x) $}
 		\end{prooftree}
 	\end{multicols}

 	Thus, $ x \in derive^{\mathcal{B}}_{I}(N, a)$ and then $ x \in derive^{\mathcal{B}}_{I}(N, A) $.
    \subsection*{Proof for Theorem \ref{th:out}: Simple-II Boolean I/O operation}
    \subsubsection*{Soundness: $out^{B}_{II}(N)$ validates $WO$ and $EQI$.}
    \begin{itemize}
 		\item [WO:] It needs to be shown that
 		
 		\begin{prooftree}
 			\AxiomC{$x \in Up(N(Eq(a))) $ } \AxiomC{$x\leq y$ }
 			\LeftLabel{WO} \BinaryInfC{$y \in Up(N(Eq(a))) $}
 		\end{prooftree}

 		If $x \in Up(N(Eq(a)))$, then there is $t_{1}$ such that $ t_{1} \leq x$ and $(a,t_{1}) \in N$ or ($(b,t_{1}) \in N$ and $a=b$). If $ x \leq y $, then $ t_{1} \leq y $ and we have $ y \in Up(N (Eq(a)))$.
 		
 		\item[EQI:] It needs to be shown that 
 		\begin{prooftree}
 			\AxiomC{$x \in  Up(N(Eq(a))) $ } \AxiomC{$a=b$ }
 			\LeftLabel{EQI} \BinaryInfC{$x \in Up(N(Eq(b)))$}
 		\end{prooftree}
 		If $ x \in  Up(N(Eq(a)))$, then there is $ t_{1} $ such that $ t_{1} \leq x $ and $ (a, t_{1}) \in N$ or ($ (c, t_{1}) \in N$ and $ a=c $). Hence, if $ a=b $, then by definition $ x \in  Up(N(Eq(b)))$.
 		
 		%then by definition of $ Eq(X) $ if $x=y$,
 		
 	\end{itemize}	
    \subsubsection*{Completeness: 	$  out^{\mathcal{B}}_{II}(N) \subseteq derive^{\mathcal{B}}_{II}(N) $.}
   It is shown that if $x \in Up(N(Eq(A))) $, then $ (A,x)  \in derive^{\mathcal{B}}_{II}(N)$. Suppose that $x \in Up(N(Eq(A))) $, then there is $ t_{1} $ such that $ t_{1} \leq x $ and $ (a, t_{1}) \in N$ or ($ (b, t_{1}) \in N$ and $ a=b $)  for $ a \in A$. There are two cases:
 	
 \begin{multicols}{2}
 		\begin{prooftree} 
 			\AxiomC{$(a, t_{1})$}
 			\AxiomC{$t_{1} \leq x $}
 			\LeftLabel{$WO$}
 			\BinaryInfC{$(a,  x)$}
 		\end{prooftree}
 		\begin{prooftree} 
 			\AxiomC{$(b, t_{1})$}
 			\AxiomC{$ a=b $}
 			\LeftLabel{$EQI$}
 			\BinaryInfC{$(a, t_{1})$}
 			\AxiomC{$t_{1} \leq x $}
 			\LeftLabel{$WO$}
 			\BinaryInfC{$(a ,  x)$}
 		\end{prooftree}
\end{multicols}
 
 	Thus, $ x \in  derive^{\mathcal{B}}_{II}(N, a) $ and then $ x \in  derive^{\mathcal{B}}_{II}(N, A)$.
 \subsection*{Proof for Theorem \ref{th:out}: Simple-minded Boolean I/O operation}
 \subsubsection*{Soundness: $ out^{\mathcal{B}}_{1} (N) $ validates $SI$ and $WO$.}
 	\begin{itemize}
 		\item[SI:] It needs to be shown that
 		
 		\begin{prooftree}
 			\AxiomC{$x \in Up(N(Up(a))) $ } \AxiomC{$b\leq a$ }
 			\LeftLabel{SI} \BinaryInfC{$x \in Up(N(Up(b))) $}
 		\end{prooftree}
 		
 		Since $ b\leq a $ we have $ Up(a) \subseteq Up(b) $. Hence, $N (Up(a)) \subseteq N(Up(b))$ and therefore $ Up(N (Up(a))) \subseteq Up(N(Up(b)) )$.	 
 		
 		\item [WO:] It needs to be show shown that
 		
 		\begin{prooftree}
 			\AxiomC{$x \in Up(N(Up(a))) $ } \AxiomC{$x\leq y$ }
 			\LeftLabel{WO} \BinaryInfC{$y \in Up(N(Up(a))) $}
 		\end{prooftree}
 		
 		Since $ Up(N(Up(a)))$ is upward-closed and $x\leq y$, we have $y \in Up(N(Up(a))) $. 
 		
 	\end{itemize}	
 \subsubsection*{Completeness: $out^{\mathcal{B}}_{1}(N) \subseteq derive^{\mathcal{B}}_{1}(N) $.}
	It is shown that if $x \in Up(N(Up(A)))$, then $ (A,x)  \in derive^{\mathcal{B}}_{1}(N)$. Suppose that $x \in Up(N(Up(A))) $, then there is $ y_{1} $ such that $ y_{1} \in N(Up(A)) $, $ y_{1} \leq x $, and there is $ t_{1} $ such that $ (t_{1}, y_{1}) \in N$ and $ a \leq t_{1} $ for $ a \in A$.
 	
 	\begin{prooftree} 
 		
 		%	\UnaryInfC{$(a,x) \quad x \vdash x \vee y$}
 		\AxiomC{$  a \leq t_{1}  $}
 		\AxiomC{$ (t_{1}, y_{1}) \quad y_{1} \leq x $ }
 		%	\UnaryInfC{$ (b, y) \quad y \vdash x \vee y $}
 		\LeftLabel{$WO$}
 		\UnaryInfC{$ (t_{1},x) $}
 		\LeftLabel{$SI$}
 		\BinaryInfC{$ (a ,  x) $}
 	\end{prooftree}
 	Thus, $ x \in  derive^{\mathcal{B}}_{1}(N, a) $ and then $ x \in   derive^{\mathcal{B}}_{1}(N, A) $.
 	
 \subsection*{Proof for Theorem \ref{th:out}: Basic Boolean I/O operation}
 \subsubsection*{Soundness: $out^{\mathcal{B}}_{2}(N)$ validates $ SI$, $WO$ and $OR$.}
 \begin{itemize}
 		\item [OR:] It needs to be shown that

 		\begin{prooftree}
 			\AxiomC{$x \in out^{\mathcal{B}}_{2}(N, \{a\}) $ } \AxiomC{$x \in out^{\mathcal{B}}_{2}(N, \{b\}) $ }
 			\LeftLabel{OR} \BinaryInfC{$x \in out^{\mathcal{B}}_{2}(N, \{a \vee b\}) $}
 		\end{prooftree}
 		
 		Suppose that $ \{ a\vee b\}  \subseteq V $, since $ V $ is saturated we have $ a \in V $ or $ b \in V $. Suppose that $ a \in V $, in this case since  $ out^{\mathcal{B}}_{2}(N, \{a\}) \subseteq  Up(N(V)) $, we have $ x \in out^{\mathcal{B}}_{2}(N, \{a \vee b\}) $.
 	\end{itemize}
 \subsubsection*{Completeness: $  out^{\mathcal{B}}_{2}(N) \subseteq derive^{\mathcal{B}}_{2}(N) $.}
  	Suppose that $ x \notin derive^{\mathcal{B}}_{2}(N, A) $, then by monotony of the derivability operation, there is a maximal set $ V $ such that $ A  \subseteq V $ and $ x \notin derive^{\mathcal{B}}_{2}(N, V) $.\footnote{Consider the set $E= \left\lbrace V: A \subseteq V \, \text{and} \, x  \notin deriv(G,V) \right\rbrace $. This set is a partially ordered set which is ordered by the monotony property of derivation. Every chain (any set linearly ordered by set-theoretic inclusion) has an upper bound (the union of the sets) in $ E $. So set $ E $ has at least a maximal element by Zorn's lemma.} $ V $ is saturated because:
 	\begin{itemize}
 		\item [(a)] Suppose that $ a \in V $ and $ a \leq b $, by definition of $ V $ we have $(a, x) \notin derive^{\mathcal{B}}_{2}(N) $.  It needs to be shown that $ x  \notin derive^{\mathcal{B}}_{2}(N,b) $ and since $ V $ is maximal, we have $ b \in V $. Suppose that $(b,x) \in derive^{\mathcal{B}}_{2}(N) $. We have
 		\begin{prooftree}
 			\AxiomC{$ (b,x) $ } \AxiomC{$ a \leq b $ }
 			\LeftLabel{SI} \BinaryInfC{$ (a,x)$}
 		\end{prooftree}
 		That is a contradiction of $(a,x) \notin derive^{\mathcal{B}}_{2}(N) $. 
 		%As a result of this item, since $ \exists z \in V\neq\emptyset $ and $z \leq  1 $ we must have $1 \in V$ as $V$ is maximal. 
 		% $  derive^{\mathcal{B}}_{2}(N,V) \neq \emptyset $ (if $ N\neq \emptyset $, then there is at least one head in $V$) there exist $ z \in derive^{\mathcal{B}}_{2}(N,V)  $ and since $z \leq  1 $ we have $1 \in V$ as $V$ is maximal.
 		%Suppose  $ x \notin derive^{\mathcal{B}}_{2}(N,V) $
 		\item [(b)] Suppose that $ a \vee b \in V$, by definition of $ V $ we have $ x \notin derive^{\mathcal{B}}_{2}(N,a\vee b) $. It needs to be shown that $ x \notin derive^{\mathcal{B}}_{2}(N,a) $ or $x \notin derive^{\mathcal{B}}_{2}(N, b) $. Suppose that $ x \in derive^{\mathcal{B}}_{2}(N,a) $ and $ x \in derive^{\mathcal{B}}_{2}(N, b) $, then we have
 		\begin{prooftree} 
 			\AxiomC{$ (a,x) $}
 			\AxiomC{$ (b, x) $ }
 			\LeftLabel{$OR$}
 			\BinaryInfC{$ (a \vee b,x) $}
 		\end{prooftree}
 		That is a contradiction of $ x \notin derive^{\mathcal{B}}_{2}(N,a\vee b) $.

 	\end{itemize}
 	
 	Therefore, we have $ x \notin Up(N( V))$ (that is equal to $x \notin  out^{\mathcal{B}}_{1}(N,V)$) and so $ x \notin out^{\mathcal{B}}_{2}(N,A) $.
 	% since  $ Up(V)=V $
 	
 \subsection*{Proof for Theorem \ref{th:out}: Reusable Boolean I/O operation} 
 \subsubsection*{Soundness: $out^{\mathcal{B}}_{3}(N)$ validates $SI$, $WO$ and $T$.}
 \begin{itemize}
 		\item [T:]  It needs to be shown that	
 		
 		\begin{prooftree}
 			\AxiomC{$x \in out^{\mathcal{B}}_{3}(N, \{a\}) $ } \AxiomC{$y \in out^{\mathcal{B}}_{3}(N, \{ x\}) $ }
 			\LeftLabel{T} \BinaryInfC{$y \in out^{\mathcal{B}}_{3}(N, \{a\}) $}
 		\end{prooftree}
 		
 		Suppose that $ X$ is the smallest set such that $ \{a\} \subseteq X=Up(X) \supseteq N(X)$. Since $x \in out^{\mathcal{B}}_{3}(N, \{a\}) $ we have $ x \in X $, and from $y \in out^{\mathcal{B}}_{3}(N, \{ x\}) $ we have $ y \in X $. Thus, $y \in out^{\mathcal{B}}_{3}(N, \{a\}) $.
 	\end{itemize}	
 \subsubsection*{Completeness: $  out^{\mathcal{B}}_{3}(N) \subseteq derive^{\mathcal{B}}_{3}(N) $.}
 	 	Suppose that $ x \notin derive^{\mathcal{B}}_{3}(N, a) $. It is necessary to find $ B $ such that $ a \in B = Up(B) \supseteq N(B) $ and $ x \notin Up(N(B)) $. 
 	Put $ B= Up(\{a\} \cup derive^{\mathcal{B}}_{3}(N, a) ) $. It is shown that $ N(B)  \subseteq B $. Suppose that $ y \in N(B) $, then there is $ b \in B $ such that $ (b,y) \in N $. It is shown that $ y \in B $. Since $ b \in B $, there are two cases:
 	\begin{itemize}
 		\item  $ b \geq a $:  in this case we have $ (a,y) \in derive^{\mathcal{B}}_{3}(N)$ since $  (b,y ) \in derive^{\mathcal{B}}_{3}(N) $ and we have
 		
 		\begin{prooftree} 
 			%	\UnaryInfC{$(a,x) \quad x \vdash x \vee y$}
 			\AxiomC{$   (b, y) $}
 			\AxiomC{$  a \leq b $ }
 			%	\UnaryInfC{$ (b, y) \quad y \vdash x \vee y $}
 			%\LeftLabel{$WO$}
 			%\UnaryInfC{$ (t_{1},x) $}
 			\LeftLabel{$SI$}
 			\BinaryInfC{$ (a ,  y) $}
 		\end{prooftree}
 		
 		\item 
 		$ \exists z \in derive^{\mathcal{B}}_{3}(N,a), b \geq z $ : in this case we have
 		
 		\begin{prooftree} 
 			%	\UnaryInfC{$(a,x) \quad x \vdash x \vee y$}
 			\AxiomC{$ (a,z)  $}
 			\AxiomC{$ (b,y) $}
 			\AxiomC{$ z\leq b $ }
 			\LeftLabel{$SI$}
 			\BinaryInfC{$ (z,y) $}
 			%		\AxiomC{$ a \wedge z \leq z$ }
 			%		\LeftLabel{$SI$}
 			%		\BinaryInfC{$ (a\wedge z,y) $}
 			\LeftLabel{$T$}
 			\BinaryInfC{$ (a ,  y) $}
 		\end{prooftree}
 		
 	\end{itemize}
 	
 	It only needs to shown that $ x \notin Up(N(B))= out^{\mathcal{B}}_{1}(N, \{a\} \cup derive^{\mathcal{B}}_{3}(N,a) )  $. Suppose that $ x \in Up(N(B)) $, then there is $ y_{1} $ such that $ x \geq y_{1} $ and $ \exists t_{1},$ $ (t_{1}, y_{1}) \in N $ and $ t_{1} \in Up(\{a\} \cup derive^{\mathcal{B}}_{3}(N, a) ) $. There are two cases:
 	\begin{itemize}
 		\item  $ t_{1} \geq a $: in this case we have 
 		
 		\begin{prooftree} 
 			%	\UnaryInfC{$(a,x) \quad x \vdash x \vee y$}
 			\AxiomC{$ (t_{1},y_{1})  $}
 			\AxiomC{$ a\leq t_{1} $}
 			\LeftLabel{$SI$}
 			\BinaryInfC{$ (a, y_{1}) $}
 			\AxiomC{$y_{1} \leq x$}
 			\LeftLabel{$WO$}
 			\BinaryInfC{$ (a, x) $}
 		\end{prooftree}
 		
 		\item 
 		$ \exists z_{1} \in  derive^{\mathcal{B}}_{3}(N,a), z_{1} \leq t_{1} $: in this case we have
 		\begin{prooftree} 
 			%	\UnaryInfC{$(a,x) \quad x \vdash x \vee y$}
 			\AxiomC{$ (a,z_{1})  $}
 			\AxiomC{$ (t_{1},y_{1}) $}
 			\AxiomC{$ z_{1} \leq t_{1} $}
 			\LeftLabel{$SI$}
 			\BinaryInfC{$ (z_{1},y_{1}) $}
 			%		\AxiomC{$ a\wedge z_{1}\leq z_{1} $ }
 			%		\LeftLabel{$SI$}
 			%		\BinaryInfC{$ (a\wedge z_{1},y_{1}) $}
 			\LeftLabel{$T$}
 			\BinaryInfC{$ (a,y_{1}) $}
 			\AxiomC{$ y_{1} \leq x $}
 			\LeftLabel{$WO$}
 			\BinaryInfC{$ (a , x) $}
 		\end{prooftree}
 		
 	\end{itemize}
 	Thus, in both cases, $ (a , x) \in derive^{\mathcal{B}}_{3}(N) $ and then $  x \in derive^{\mathcal{B}}_{3}(N, a) $, and that is a contradiction.

 \subsection*{Proof for Theorem \ref{ob : th} } 	
 		The proof is based on the reversibility of inference rules, as studied by Makinson and van der Torre~\cite{Makinson00}.
 	% In the proof systems that we just have AND, SI, WO, OR, CT rules; by putting AND only at the end, we don't lose any derivation from our system.   
 	\begin{lemma}
 		Let $ D $ be any derivation using at most EQI, SI, WO, OR, AND, CT. Then, there is a derivation $ D^{'} $ of the same root from a subset of leaves that applies AND only at the end.
 	\end{lemma}
 	
 	\begin{proof}
 		See Observation 18~\cite{Makinson00}.
 		
 		The main point of the observation is that it is possible to reverse the order of rules AND, WO to WO, AND; AND, SI to SI, AND; AND, OR to OR, AND and finally AND, CT to SI, CT or CT, AND. It is also possible to reverse the order of the AND and EQI rules as follows:
 		\begin{multicols}{2}	
 			\begin{prooftree}\hspace*{-.4in} 
 				\AxiomC{$(a,x)$} \AxiomC{$(a,y)$}
 				\LeftLabel{AND} \BinaryInfC{$(a,x\wedge y)$}
 				\AxiomC{$ a=b $}
 				\LeftLabel{EQI}
 				\BinaryInfC{$(b, x\wedge y)$}
 			\end{prooftree}
 			\begin{prooftree}\hspace*{-.5in}
 				\AxiomC{$(a,x) $ $a=b $ }  
 				\LeftLabel{EQI} \UnaryInfC{$(b,x)$}
 				\AxiomC{$(a,y)$ $a=b$}  
 				\LeftLabel{EQI} \UnaryInfC{$(b,y)$}
 				\LeftLabel{AND}
 				\BinaryInfC{$(b,x\wedge y)$}
 			\end{prooftree}
 		\end{multicols}
 	\end{proof}
 	
 	Hence, in each system of $\{WO, EQI, AND\}$, $\{SI, WO, AND\}$ and $\{SI,$ $ WO, OR, AND\}$,  the AND rule can be applied only at the end. Thus, it is possible to characterize $ deriv^{AND}_{i}(N) $ using the fact $ deriv^{\mathcal{B}}_{i}(N) = out_{i}^{\mathcal{B}} (N)$ and the iterations of AND.
 	%the such that is closed under iteration of AND rule. 
 	
 	It is easy to check that CT can be reversed with SI, EQO, WO, and EQI by the fact that it is similarly possible to characterize $ deriv^{CT}_{i}(N) $. 
 	
 	Finally, since AND can be reversed with ST, WO and CT, it is possible to characterize $deriv^{CT, AND}_{1}(N)$ by applying (finite) iterations of AND over $out_{1}^{CT}(N)$ that means $out_{1}^{CT, AND}(N)$.

 	\subsection*{Proof for Theorem \ref{Ab: th} } 
  
  The proofs are the same as the soundness and completeness proofs in Theorem~\ref{th:out}.

 	\subsection*{Proof for Theorem \ref{Nes} } 
 		This only looks at I/O operations over Boolean algebras since the argument for abstract logics is similar. It needs to be shown that 
	\begin{itemize}
		\item $ N \subseteq out_{i}^{\mathcal{B}}(N)  $
		\item $ N \subseteq M \Rightarrow out_{i}^{\mathcal{B}}(N) \subseteq out_{i}^{\mathcal{B}}(M) $
		\item $ out_{i}^{\mathcal{B}}(N)  = out_{i}^{\mathcal{B}} (out_{i}^{\mathcal{B}}(N))  $
	\end{itemize} 

By the soundness and completeness theorems, we have  $ out_{i}^{\mathcal{B}}(N) =$ \linebreak $ derive_{i}^{\mathcal{B}} (N) $. So $ derive_{i}^{\mathcal{B}} (N) $ is studied, which is more simple than $ out_{i}^{\mathcal{B}} (N)$. The first two properties are clear from the  definition of $ derive_{i}^{\mathcal{B}} $.  For the last property, it needs to be shown that $ derive_{i}^{\mathcal{B}} (N) = derive_{i}^{\mathcal{B}} (derive_{i}^{\mathcal{B}} (N)) $. We have $   derive_{i}^{\mathcal{B}} (derive_{i}^{\mathcal{B}} (N)) = derive_{i}^{\mathcal{B}} ( \{ (A, x) | (a,x) \in derive_{i}^{\mathcal{B}}(N) $  $ \text{for some}$ $a \in A\}) = \{ (A, x) | (a,x) \in derive_{i}^{\mathcal{B}}(N) $  $ \text{ for some } a \in A\} $ since $ N \subseteq \{ (A, x) | $ $ (a,x) \in derive_{i}^{\mathcal{B}}(N) $ for some  $a \in A\}  $ and the same rules apply over $ derive_{i}^{\mathcal{B}}(N) $. 
	Actually, it needs to be shown that if $ (a,x) \in derive_{i}^{\mathcal{B}}(N) $, then $  derive_{i}^{\mathcal{B}}(N) = derive_{i}^{\mathcal{B}}(N\cup \{(a,x) \} ) $  holds for $ derive_{i}^{\mathcal{B}} $.
\section*{Appendix C} \label{sec:con} 

\subsection*{Proof for Theorem \ref{th: Bo} } 
 See~\cite{Blok,jansana2016algebraic}. 
 
\subsection*{Proof for Theorem \ref{th : Fm} } 
	Here is the proof for the case of $ i=1 $.  
	\begin{itemize}
		\item Suppose that $ (\varphi,\psi) \in derive^{\mathbf{Fm}(X)}_{1}(N)$. For an arbitrary  valuation $ V $ and arbitrary Boolean algebra $ \mathcal{B} \in \mathbf{BA} $, it needs to be shown that $ V(\psi) \in out^{\mathcal{B}}_{1} (N^{V}, \{ V(\varphi)\})$. The proof is by induction on the length of the proof $(\varphi, \psi) \in derive^{\mathbf{Fm}(X)}_{1}(N) $.
		\\
		\textit{Base case:} If $(\varphi, \psi) \in  N$, then $ (V(\varphi), V(\psi)) \in N^{V} $ by definition, and we have $ V(\psi) \in out^{\mathcal{B}}_{1} (N^{V},  \{ V(\varphi)\})$.
		\\
		\textit{Inductive step:} It is shown that for $ n > 0 $, if $ V(\psi) \in out^{\mathcal{B}}_{1} (N^{V},  \{ V(\varphi)\})$ holds for $ n $, then $ V(\psi) \in out^{\mathcal{B}}_{1} (N^{V},  \{ V(\varphi)\})$ also holds for $ n +1 $.
		
		Suppose that the length of proof $(\varphi,\psi) \in derive^{\mathbf{Fm}(X)}_{1}(N) $ is $ n + 1 $. There are two possibilities:
		\begin{itemize}
			\item \textit{Using SI in the last step:}
			There is $ \phi $ such that $ (\phi, \psi) \in $ \linebreak $ derive^{\mathbf{Fm}(X)}_{1}(N) $ and $ \varphi \vdash_{C} \phi$. In this case, by the induction step we have $ V(\psi) \in out^{\mathcal{B}}_{1} (N^{V},  \{ V(\phi)\})$, and by the completeness of the  simple-minded operation we have  $ (V(\phi), V(\psi)) \in derive^{\mathcal{B}}_{1}(N) $. Since $ \varphi \vdash_{C} \phi$, then by Theorem~\ref{th: Bo} we have $ \varphi \vDash_{\mathbf{BA}} \phi $. So \linebreak $ V(\varphi) \wedge V(\phi) = V(\varphi) $. Then from  $ (V(\phi), V(\psi)) \in derive^{\mathcal{B}}_{1}(N) $ and $ V(\varphi) \leq V(\phi) $ using the SI rule we have $ (V(\varphi), V(\psi)) \in derive^{\mathcal{B}}_{1}(N) $, and by the soundness of the simple-minded operation we have $ V(\psi) \in out^{B}_{1} (N^{V},  \{ V(\varphi)\})$.  
			\item \textit{Using WO in the last step:}
			There is $ \phi $ such that $(\varphi, \phi) \in $ \linebreak $derive^{\mathbf{Fm}(X)}_{1}(N) $ and $ \phi \vdash_{C} \psi$. In this case, by the induction step we have $ V(\phi) \in out^{\mathcal{B}}_{1} (N^{V},  \{ V(\varphi)\})$, and by the completeness of the simple-minded operation we have $ (V(\varphi), V(\phi)) \in derive^{\mathcal{B}}_{1}(N) $. Since $ \phi \vdash_{C} \psi$, then by Theorem~\ref{th: Bo} we have $ \phi \vDash_{\mathbf{BA}} \psi $. So \linebreak $ V(\phi) \wedge V(\psi) = V(\phi) $. Then from  $ (V(\varphi), V(\phi)) \in derive^{\mathcal{B}}_{1}(N) $ and $ V(\phi) \leq V(\psi) $ using the WO rule we have $ (V(\varphi), V(\psi)) \in  derive^{\mathcal{B}}_{1}(N) $, and by the soundness of the simple-minded operation we have $ V(\psi) \in   out^{\mathcal{B}}_{1} (N^{V}, \{ V(\varphi)\})$. 
		\end{itemize} 
		
		\item The proof in the other direction is by contraposition. Suppose  that $(\varphi,\psi)\notin$  $  derive^{\mathbf{Fm}(X)}_{1}(N)$, if $ \mathbf{Fm}(X)$ is taken as a Boolean algebra,  then by the completeness of $ derive^{\mathbf{Fm}(X)}_{1}(N) $, we have $ \psi \notin out^{\mathbf{Fm}(X)}_{1} (N, \{\varphi\})  $. Then  it is enough that the valuation function is put as the identity function on the Boolean algebra $ Fm(X) $ which means\\ $ \psi \notin out^{\mathcal{B}=\mathbf{Fm}(X)}_{1} (N, \{\varphi\})  $.  
	\end{itemize}
	The proof is similar for the other derivation systems: $derive^{\mathbf{Fm}(X)}_{R}(N)$, \\$  derive^{\mathbf{Fm}(X)}_{L}(N)$, $derive^{\mathbf{Fm}(X)}_{I}(N)$, $  derive^{\mathbf{Fm}(X)}_{II}(N)$, $derive^{\mathbf{Fm}(X)}_{2}(N)$, and \\$ derive^{Fm(X)}_{3}(N)$.

\subsection*{Proof for Theorem \ref{th : AND} } 	
\begin{itemize}
		\item  The proof from right to left is similar to Theorem~\ref{th : Fm}. It just needs to be checked that for the case when AND  is the last step of the derivation, that there are $ \delta_{1} $ and $ \delta_{2} $ such that $ (\varphi, \delta_{1}), (\varphi, \delta_{2})  \in derive^{AND}_{i}(N)$ and $ \psi = \delta_{1} \wedge \delta_{2} $. In this case, by the induction step we have $V(\delta_{1}) \in out^{AND}_{i} (N^{V},  \{ V(\varphi)\})$ and $V(\delta_{2}) \in out^{AND}_{i} (N^{V},  \{ V(\varphi)\})$. By Theorem~\ref{ob : th}, we have $ (\varphi, \delta_{1}) \in derive^{AND}_{i}(N) $  and $ (\varphi, \delta_{2}) \in derive^{AND}_{i}(N) $. Then by using the AND rule, we have  $ (\varphi, \delta_{1} \wedge \delta_{2}) \in derive^{AND}_{i}(N) $, and then by Theorem \ref{ob : th}, we have $ V(\psi) \in out^{AND}_{i} (N^{V},  \{ V(\varphi)\})$. 
		
		\item The proof in the other direction is by contraposition. Suppose that $(\varphi, \psi) \notin $  $derive^{AND}_{1}(N) $, if $ \mathbf{Fm}(X) $ is taken as a Boolean algebra,  then by Theorem~\ref{ob : th}, we have $ \psi \notin out^{AND}_{1} (N, \{\varphi\})  $, then if the valuation function is put as the identity function on the algebra $ Fm(X) $, we have $ \psi \notin out^{AND}_{1} (N, \{\varphi\})$.  
	\end{itemize}
 It is possible to extend the proof for the arbitrary input set $ \varGamma \subseteq Fm(X) $ and to extend this theorem for other addition rule operators. 
 
\subsection*{Proof for Theorem \ref{th : con} }  
\begin{itemize}
		\item From left to right:
		Suppose that $ (\varphi,\psi) \in derive^{Con}_{i}(N) $, then by definition,  $ (\varphi,\psi) \in derive^{\mathbf{Fm}(X)}_{i}(N) $ and $ Con, \psi \nvdash_{C} \perp  $. 
		From Theorem~\ref{th : Fm} we have ``$V(\psi) \in out^{\mathcal{B}}_{i}$  $(N^{V},$ $ \{ V(\varphi)\})$ for every $ \mathcal{B} \in \mathbf{BA} $ and valuation  $V$'', and from Theorem~\ref{th: Bo} there is a  Boolean algebra $ \mathcal{B} $ such that $ Con , \psi \nvDash_{\mathcal{B}} \perp $. So there is a valuation $ V $ on $ \mathcal{B} $  such that $\forall \delta\in Con, \, V(\delta \wedge \psi )= 1_{\mathcal{B}}$.
		\item  The proof from right to left is similar. 
	\end{itemize}
	By the definition of $ derive^{Con}_{i}(N) $, it is possible to extend the theorem for the case of $ (\varGamma,\psi) \in derive^{Con}_{i}(N) $  where $ \varGamma \subseteq Fm(X) $.
 
 \subsection*{Proof for Theorem \ref{th : OH} } 
 \begin{itemize}
		\item From left to right: Suppose that $ \varphi> \bigcirc \psi \in derive^{O^{H}}_{i}(N)$,  by definition, $ (\varphi,\psi) \in derive^{\mathbf{Fm}(X)}_{i}(N) $ and from Theorem~\ref{th : Fm}, we have ``$V(\psi) \in out^{\mathcal{B}}_{i} (N^{V},  \{ V(\varphi)\})$ for every $ \mathcal{B} \in \mathbf{BA} $ and valuation  $V$''. For the second part, notice that every maximal consistent subset defines a valuation and vice versa. So ``$\forall M \in \opf (\varphi) (\psi \in M)$'' is equivalent to that  for any valuation $V_{i} \in opt_{\succeq_{f}} (\varphi) $, so that we have $V_{i}(\psi)= 1_{\mathcal{B}}$ and vice versa.
		\item From right to left, the proof is similar. 
	\end{itemize}
	By the definition of $ derive^{O^{H}}_{i}(N) $, it is possible to extend the theorem for the case of	$ \varGamma > \bigcirc\psi \in derive^{O^{H}}_{i}(N) $  where $ \varGamma \subseteq Fm(X) $.
	
\subsection*{Proof for Theorem \ref{th : OK} } 	
	\begin{itemize}
		\item From left to right: Suppose that $ (\varphi,\psi) \in derive^{O^{K}}_{i}(N)$,  by definition, $ (\varphi,\psi) \in derive^{\mathbf{Fm}(X)}_{i}(N)$ and from Theorem~\ref{th : Fm} we have ``$V(\psi) \in out^{B}_{i} (N^{V},  \{ V(\varphi)\})$ for every $ \mathcal{B} \in \mathbf{BA} $ and valuation  $V$''. For the second part, notice that every maximal consistent subset defines a valuation and vice versa. So ``$\forall M \in opt_{f^{A}} (\varphi) (\psi \in M)$'' is equivalent to that for any valuation $V_{i} \in opt_{\succeq_{A}} (\varphi) $, so that we have $ V_{i}(\psi)= 1_{\mathcal{B}}$ and vice versa.
		\item From right to left, the proof is similar. 
	\end{itemize}
	By the definition of $ derive^{O^{K}}_{i}(N) $, it is possible to extend the theorem for the case of	$ \varGamma > \bigcirc \psi \in derive^{O^{K}}_{i}(N) $  where $ \varGamma \subseteq Fm(X) $.
\section*{Appendix D} \label{sec:hol} 
\subsection*{Introduction to higher-order logic}
This brief introduction to classical higher-order logic (HOL) is adapted from~\cite{J46}. 
 
HOL is based on simple typed $ \lambda $-calculus. It is assumed that the set  $ \mathcal{T} $ of simple types is freely generated from a set of basic types $ \{o,i \} $ using the function type constructor $ \rightarrow $. Type $ o $ denotes the set of Booleans where type $i$ refers to a non-empty set of individuals.

For $ \alpha, \beta, o \in \mathcal{T}$, the \emph{language of HOL} is generated as follows:
 
\[
\begin{array}{lcl}
& s,t ::= p_{\alpha} | X_{\alpha}|  (\lambda X_{\alpha}s_{\beta})_{\alpha \rightarrow \beta}| (s_{\alpha \rightarrow \beta}\, t_\alpha)_{\beta} % | \\
\end{array}
\]
where $ p_{\alpha}$ represents a typed constant symbol (from a possibly infinite set $\mathcal{P}_\alpha$ of such constant symbols) and $ X_{\alpha}$ represents a typed variable symbol (from a possibly infinite set $\mathcal{V}_\alpha$ of such symbols). 
%We assume the sets of typed constant symbols and the set of typed variable symbols 
$(\lambda X_{\alpha}s_{\beta})_{\alpha \rightarrow \beta}$  and $(s_{\alpha \rightarrow \beta}\, t_\alpha)_{\beta}  $ are called \textit{abstraction} and \textit{application} respectively. HOL is a logic of terms in the sense that the \emph{formulas of HOL} are given as terms of type $o$. Moreover, a sufficient number of primitive logical connectives are required in the signature of HOL, i.e., these logical connectives must be contained in the sets $\mathcal{P}_\alpha$ of constant symbols.
The primitive logical connectives of choice in this article are $ \neg_{o \rightarrow o}  $, $\vee_{o \rightarrow o \rightarrow o}  $, $\Pi_{(\alpha \rightarrow o)\rightarrow o} $ and $=_{\alpha \rightarrow \alpha \rightarrow o}$. The symbols $\Pi_{(\alpha \rightarrow o)\rightarrow o} $ and $=_{\alpha \rightarrow \alpha \rightarrow o}$ are generally assumed for each type $ \alpha \in \mathcal{T} $. From the selected set of primitive connectives, other logical connectives can be introduced as abbreviations. Type information as well as brackets may be omitted if obvious from the context, and infix notation may also be used to improve readability. For example, $(s\vee t) $ may be written instead of $((\vee_{o \rightarrow o \rightarrow o} \, s_{o})\, t_{o})_{o}$. Often,  $\forall X_\alpha s_o$ is written as syntactic sugar for $\Pi_{(\alpha \rightarrow o)\rightarrow o} (\lambda X_\alpha s_o)$.
% \footnote{Type information as well as brackets may be omitted if obvious from the context, and we may also use infix notation to improve readability. For example, we may write $(s\vee t) $ instead of $((\vee_{o \rightarrow o \rightarrow o} \, s_{o})\, t_{o})_{o}$. }

The notions of \emph{free variables}, \emph{$\alpha$-conversion},
\emph{$\beta\eta$-equality} and
\emph{substitution} of a term $s_{\alpha}$ for a variable $X_{\alpha}$
in a term $t_{\beta}$, denoted as $[s/X]t$, are defined as usual. 
%, denoted as $=_{\beta\eta}$,

The semantics of HOL are well understood and thoroughly documented~\cite{J6}. In this article, the semantics of choice is Henkin's general models~\cite{Henkin50}. 
%For more details refer to the literature \cite{J6}.

A \emph{frame} $D$ is a collection $\{D_\alpha\}_{\alpha\in \mathcal{T}}$ of nonempty sets $D_\alpha$ such that $D_o = \{T,F\}$, denoting truth and falsehood respectively.  $D_{\alpha\rightarrow\beta}$ represents a collection of functions mapping $D_\alpha$ into $D_\beta$. 

A \emph{model} for HOL is a tuple $M= \langle D, I \rangle$, where $D$ is a frame and
$I$ is a family of typed interpretation functions mapping constant symbols $ p_\alpha $ to appropriate elements of $D_\alpha$ called the \emph{denotation of $ p_\alpha$}. The logical connectives $\neg$, $\vee$, $\Pi$ and $=$ are always given in their expected standard denotations. A \emph{variable assignment} $g$ maps variables $X_\alpha$ to elements in $D_\alpha$ while $g[d/W]$ denotes the assignment that is identical to $g$, except for the variable $W$, which is now mapped to $d$.  The \emph{denotation} $ \| s_\alpha\|^{M,g}$ of a HOL term
$ s_\alpha$ on a model $ M= \langle D, I \rangle$ under assignment $g$ is an element $d\in D_\alpha$ and is defined in the following way: 
\begin{center}
	\begin{tabular}{rcl}
		$ \|p_\alpha\|^{M,g}$ & $=$ & $ I(p_\alpha)$ \\[.5em]
		$ \|X_\alpha\|^{M,g}$ & $=$ & $ g(X_\alpha)$ \\[.5em]
		$ \|(s_{\alpha\rightarrow\beta}\, t_\alpha)_\beta\|^{M,g} $ & $=$ &
		$ \|s_{\alpha\rightarrow\beta}\|^{M,g}(\|t_\alpha\|^{M,g}) $ \\[.5em]
		$ \|(\lambda{X_\alpha} s_\beta)_{\alpha\rightarrow\beta}\|^{M,g} $ & $= $ &
		the function $ f $ from $ D_\alpha $ to $ D_\beta $ such
		that \\
		& & $ f(d)  =  \|s_\beta\|^{M,g[d/X_\alpha]}$ for all
		$ d\in  D_\alpha$
	\end{tabular}
\end{center}

Since $ I(\neg_{o \rightarrow o}) $, $ I(\vee_{o \rightarrow o \rightarrow o}) $, $ I(\Pi_{(\alpha \rightarrow o)\rightarrow o}) $ and $ I(=_{\alpha \rightarrow \alpha \rightarrow o}) $ always denote the standard truth functions, we have:

\begin{enumerate}
	\item $ \|(\neg_{o\rightarrow o}\, s_o)_o\|^{M,g}   =  T $ \quad  iff \quad
	$ \|s_o\|^{M,g}  =  F $. 
	\item $ \|(({\vee_{o\rightarrow o\rightarrow o}}\,s_o)\,t_o)_o\|^{M,g}  =
	T $\quad  iff \quad $ \|s_o\|^{M,g}  =  T $ or $ \|t_o\|^{M,g} 
	=  T $. 
	\item $ \|(\forall X_\alpha s_o)_o\|^{M,g}  
	=  \|(\Pi_{(\alpha \rightarrow o)\rightarrow o}(\lambda{X_\alpha} s_o))_o\|^{M,g} 
	=  T $\quad  iff \quad  for all $d\in
	D_\alpha $ we have $ \|s_o\|^{M,g[d/X_\alpha]}= T$.
	\item  $ \| ((=_{\alpha \rightarrow \alpha \rightarrow o} \, s_{\alpha} ) \, t_\alpha)_{o}\|^{M,g} =  T $ \quad  iff \quad
	$ \|s_\alpha\|^{M,g}  = \|t_\alpha\|^{M,g} $.
\end{enumerate}
From the selected set of primitive connectives, other logical connectives can be introduced as abbreviations. An HOL formula $s_o$ is \emph{true} in a Henkin model $M$ under the assignment $g$ if and only if $\|s_o\|^{M,g} = T$.
This is also expressed with the notation $M,g \models^\text{\itshape HOL} s_o$.  An HOL formula $s_o$ is called \emph{valid} in $M$, denoted as $M\models^\text{\itshape HOL} s_o$, if and only if
$M,g \models^\text{\itshape HOL} s_o$ for all assignments $g$. Moreover, a
formula $s_o$ is called \emph{valid}, denoted as
$\models^\text{\itshape HOL} s_o$, if and only if $s_o$ is valid in all
Henkin models $M$. Finally,  $\Sigma\models^\text{\itshape HOL} s_o$ is defined for a set of HOL
formulas $\Sigma$ if and only if $M\models^\text{\itshape HOL} s_o$ for
all Henkin models $M$
with $M\models^\text{\itshape HOL} t_o$ for all $t_o\in \Sigma$.

\subsection*{Proof for Lemma \ref{lemma:1}}
	The proof is straightforward. For example, for  COM $ \vee $ we have the following:
 \linebreak\linebreak\linebreak\linebreak
	\noindent COM $ \vee $: \\[-1.75em]
	\begin{tabbing}
		\qquad \= 
		For all $a, b \in D_{i}$: $I \vee_{i\typearrow i \typearrow i} \, a \, b = I \vee_{i\typearrow i \typearrow i} \, b \, a $  \\ 
		\>  (from the definition of $  I \vee_{i\typearrow i \typearrow i}$ and $ \vee $ ) \\
		%boolean operator
		$\Leftrightarrow$ 
		\>  For all  assignments $g$, for all $ a, b \in D_{i}$ \\
		\> $\| X \vee Y = Y \vee X \|^{{H}^{M},g[a/X_{i}][b/Y_{i}] }=T$ \\
		
		$\Leftrightarrow$ 
		\> For all $g$,   we have $\| \forall X \forall Y (X \vee Y = Y \vee X )\|^{{H}^{\mathcal{N}},g  }=T$ \\
		
		$\Leftrightarrow$ 
		\> ${H}^{\mathcal{N}} \models^\text{HOL} \text{COM}  \vee  $
	\end{tabbing}

\subsection*{Proof for Lemma \ref{lemmaab:3}}
 	
	Fact: notice that for all $\varphi \in \mathbf{Fm}(X) $ and for all assignments $g$ by induction on the structure of  $ \varphi$, we have $\|\lfloor \varphi\rfloor  \|^{{H}^{\mathcal{N}}, g } = V(\varphi)$. %translates element
 
	For simplification, the term abbreviations are used for  the saturated set, the $ \leq $ ordering and upward set. It is easy to see that these terms abbreviations have the same corresponding sets in the corresponding Henkin model as in the Boolean algebra. 
	
	Here then is the proof:
 \\

	\noindent ($d_{1}(N)$) 
	% (b) $\delta = \neg \varphi$. In this case: \\[-1.5em]
	\begin{tabbing}
		\qquad \= 
		$ \| \lfloor d_{1}(N) (\varphi,\psi) \rfloor \|^{{H}^{\mathcal{N}}, g} = T  $\\
		
		$\Leftrightarrow$ 
		\> $ \|  (\bigcirc_{1}(N)_{\proptype \typearrow \proptype} \{\lfloor \varphi\rfloor \})  \lfloor \psi  \rfloor \|^{{H}^{\mathcal{N}}, g} = T$\\
		
		$\Leftrightarrow$ 
		\> $ \| ( \lambda A_\proptype \lambda X_\itype (\exists U \,  ( \exists Y \, (\exists Z \, (A \, Z \wedge Z \leq Y $
		\\
		\>$\wedge N \, Y \, U \wedge U \leq X )   ) ) ) \{\lfloor \varphi\rfloor \} ) \lfloor \psi  \rfloor \|^{{H}^{\mathcal{N}}, g} = T$\\
		
		$\Leftrightarrow$ 
		\> $ \| (\lambda X_\itype (\exists U \,  ( \exists Y \, (\exists Z \, (\{\lfloor \varphi\rfloor \} \, Z \wedge Z \leq Y $ \\
		\>$\wedge N \, Y \, U \wedge U \leq X )  )) ) )\lfloor \psi  \rfloor \|^{{H}^{\mathcal{N}}, g} = T$\\
		 
		 $\Leftrightarrow$ 
		 \> $ \| \exists U \,  ( \exists Y \, (\exists Z \, (\{\lfloor \varphi\rfloor \} \, Z \wedge Z \leq Y \wedge N \, Y \, U \wedge U \leq \lfloor \psi  \rfloor )  ) ) \|^{{H}^{\mathcal{N}}, g} = T$\\
		 
		 $\Leftrightarrow$ 
		 \> $ \| \exists U \,  ( \exists Y \, (  \, \lfloor \varphi\rfloor \leq Y \wedge N \, Y \, U \wedge U \leq \lfloor \psi  \rfloor  )) \|^{{H}^{\mathcal{N}}, g} = T$\\
		
		$\Leftrightarrow$ 
		\>There are elements $ b $ and $ c $ such that $ b, c \in D_i $ and 
		\\
		\> $ \| \lfloor \varphi\rfloor \leq Y \wedge N \, Y \, U \wedge U \leq \lfloor \psi\rfloor\|^{{H}^{M}, g [b /U_\itype] [c /Y_\itype]    } = T$\\
		
		$\Leftrightarrow$ 
		\>There are elements $ b, c \in B $ such that
		\\
		\> $     V (\varphi) \leq c \wedge N^V \, c \, b \wedge b \leq   V(\psi)   $\\
		
		$\Leftrightarrow$ 
		\>  $ V(\psi) \in Up( N^V (Up (\{V(\varphi)\}))) $ \\
		
		$\Leftrightarrow$ 
		\>  $V(\psi) \in out^{\mathcal{B}}_{1} (N^V, \{V(\varphi)\})  $ \\
		
	\end{tabbing}	
	%$ \| \lfloor \varphi\rfloor S\|^{{H}^{M}, g[s/S_\itype]} = F $  
	\pagebreak
	\noindent ($d_{2}(N)$)
	% (b) $\delta = \neg \varphi$. In this case: \\[-1.5em]
	\begin{tabbing}
		\qquad \= 
		$ \| \lfloor d_{2}(N) (\varphi,\psi) \rfloor \|^{{H}^{\mathcal{N}}, g} = T  $\\
		
		$\Leftrightarrow$ 
		\> $ \|  (\bigcirc_{2}(N)_{\proptype \typearrow \proptype} \{\lfloor \varphi\rfloor \} ) \lfloor \psi \rfloor \|^{{H}^{\mathcal{N}}, g} = T$\\
		
		$\Leftrightarrow$ 
		\> $ \| ( \lambda A_\proptype \lambda X_\itype (\forall V \, ( Saturated \, V \wedge \forall U ( A \, U \rightarrow V \, U)$ 
		\\
		\>$\rightarrow \exists Y \, ( \exists Z \, (Z \leq X \wedge N \, Y \, Z \wedge V \, Y) ) ) ) \{\lfloor \varphi\rfloor \} ) \lfloor \psi  \rfloor \|^{{H}^{\mathcal{N}}, g} = T$\\
		
		$\Leftrightarrow$ 
		\> $ \|   ( \lambda X_\itype (\forall V \, ( Saturated \, V \wedge \forall U ( \{\lfloor \varphi\rfloor \}  \, U \rightarrow V \, U)$ 
		\\
		\>$\rightarrow \exists Y \, ( \exists Z \, (Z \leq X \wedge N \, Y \, Z \wedge V \, Y) ) ) ) ) \lfloor \psi  \rfloor \|^{{H}^{\mathcal{N}}, g} = T$\\
		
		$\Leftrightarrow$ 
		\> $ \|   \forall V \, ( Saturated \, V \wedge \forall U ( \{\lfloor \varphi\rfloor \}  \, U \rightarrow V \, U)$ 
		\\
		\>$\rightarrow \exists Y \, ( \exists Z \, (Z \leq \lfloor \psi  \rfloor \wedge N \, Y \, Z \wedge V \, Y) )  )   \|^{{H}^{\mathcal{N}}, g} = T$\\
		
	    $\Leftrightarrow$ 
	    \>There are elements $ b $ and $ c $ such that $ b, c \in D_i $ and \\
		\> $ \| \forall V \, ( Saturated \, V \wedge \forall U ( \{\lfloor \varphi \rfloor \}  \, U \rightarrow V \, U)$ 
		\\
		\>$\rightarrow   (Z \leq \lfloor \psi  \rfloor \wedge N \, Y \, Z \wedge V \, Y) )    \|^{{H}^{\mathcal{N}}, g[b /Y_\itype] [c /Z_\itype]} = T$\\

		$\Leftrightarrow$ 
		\>For every saturated set $ V $ that $ \{V(\varphi)\} \subseteq V $, 
		\\
		\> there are elements $ b, c \in B $ such that 
		\\
		\>   $c \leq  V(\psi)  \wedge N^V \, b \, c \wedge V \, b  $\\
		
		$\Leftrightarrow$ 
		\> For every saturated set $ V $ such that $ \{V(\varphi) \} \subseteq V $, \\
		\> we have $ V(\psi) \in Up( N^V (V))) $ \\
		
		$\Leftrightarrow$ 
		\>  $V(\psi) \in out^{\mathcal{B}}_{2} (N^V, \{V(\varphi)\})  $ \\
		
	\end{tabbing}	
	\noindent ($d_{3}(N)$)
	% (b) $\delta = \neg \varphi$. In this case: \\[-1.5em]
	\begin{tabbing}
		\qquad \= 
		$ \| \lfloor d_{3}(N) (\varphi,\psi) \rfloor \|^{{H}^{\mathcal{N}}, g} = T  $\\
		
		$\Leftrightarrow$ 
		\> $ \|  (\bigcirc_{3}(N)_{\proptype \typearrow \proptype} \{\lfloor \varphi\rfloor \})  \lfloor \psi \rfloor \|^{{H}^{\mathcal{N}}, g} = T$\\
		
		$\Leftrightarrow$ 
		\> $ \|  (\lambda A_\proptype \lambda X_\itype (\forall V ( \forall U ( A \, U \rightarrow V \, U) \wedge V = Up \, V $ 
		\\
		\>	$ \wedge \forall W (\exists Y (V \, Y \wedge N \, Y \, W) \rightarrow  V \, W  )    $
		\\
		\>$\rightarrow \exists Y \, ( \exists Z \, (Z \leq X \wedge N \, Y \, Z \wedge V \, Y ) ) ) ) \{\lfloor \varphi\rfloor \} ) \lfloor \psi  \rfloor \|^{{H}^{\mathcal{N}}, g} = T$\\
		
		$\Leftrightarrow$ 
		\> $ \|   ( \lambda X_\itype (\forall V ( \forall U (  \{\lfloor \varphi\rfloor \} \, U \rightarrow V \, U) \wedge V = Up \, V $ 
		\\
		\>	$ \wedge \forall W (\exists Y (V \, Y \wedge N \, Y  \, W) \rightarrow  V \, W  )    $
		\\
		\>$\rightarrow \exists Y \, ( \exists Z \, (Z \leq X \wedge N \, Y \, Z \wedge V \, Y ) ) ) ) ) \lfloor \psi  \rfloor \|^{{H}^{\mathcal{N}}, g} = T$\\
		
		$\Leftrightarrow$ 
		\> $ \|     \forall V ( \forall U (  \{\lfloor \varphi\rfloor \} \, U \rightarrow V \, U) \wedge V = Up \, V $ 
		\\
		\>	$ \wedge \forall W (\exists Y (V \, Y \wedge N \, Y \, W) \rightarrow  V \, W  )   $
		\\
		\>$\rightarrow \exists Y \, ( \exists Z \, (Z \leq \lfloor \psi  \rfloor \wedge N \, Y \, Z \wedge V \, Y ) ) ) \|^{{H}^{\mathcal{N}}, g} = T$\\
		
		$\Leftrightarrow$ 
		\> There are elements $ b $ and $ c $ such that $ b, c \in D_i $ and 
		\\
		\> $ \|     \forall V ( \forall U (  \{\lfloor \varphi\rfloor \} \, U \rightarrow V \, U) \wedge V = Up \, V $ 
		\\
		\>	$ \wedge \forall W (\exists Y (V \, Y \wedge N \, Y \, W) \rightarrow  V \, W  )   $
		\\
		\>$\rightarrow   (Z \leq \lfloor \psi  \rfloor \wedge N \, Y \, Z \wedge V \, Y ))  \|^{{H}^{\mathcal{N}}, g [b /Y_\itype] [c /Z_\itype] } = T$\\

		$\Leftrightarrow$ 
		\>For every set $ V $ that $ Up(V)=V $, $ \{ V(\varphi) \} \subseteq V $ and $ N^V (V) \subseteq V $,
		\\
		\>there are elements $ b, c \in B $ such that     
		\\
		\>   $     c \leq  V(\psi)  \wedge N^V \, b \, c \wedge V \, b    $\\
		
		$\Leftrightarrow$ 
		\> For every set $ V $ that $ Up(V)=V $, $ \{ V(\varphi) \} \subseteq V $ and $ N^V (V) \subseteq V $,\\
		\> we have $ V(\psi) \in Up( N^V (V))) $ \\
		
		$\Leftrightarrow$ 
		\>  $V(\psi) \in out^{\mathcal{B}}_{3} (N, \{V(\varphi)\})  $ \\
		
	\end{tabbing}	
	
 \subsection*{Proof for Lemma \ref{lemmaab4}}
 % Throughout the proof  whenever possible we omit types in order to avoid making the notation too cumbersome. 
	Suppose that
	${H} = \langle \{D_\alpha\}_{\alpha \in {T}}, I \rangle$ is a
	Henkin model such that
	${H} \models^\text{HOL} \Sigma$ for all $\Sigma\in\{COM \vee,..., Dis \wedge \vee \} $. Without loss of generality, it can
	be assumed that the domains of $H$ are denumerable~\cite{Henkin50}.
	The corresponding Boolean normative model $ \mathcal{N}$ is constructed as follows:
	
	\begin{itemize}[topsep=1pt,itemsep=0ex,partopsep=1ex,parsep=1ex]
		\item
		$ B= D_{i} $.  
		\item 
		$ 1 = I \top_{i} $.
		\item 
		$ 0 = I \bot_{i} $.
		\item 
		$ a \vee b = c $  for $ a,b, c \in B $ iff $I \vee_{i\typearrow i\typearrow i } a b = c$.
		\item
		$ a \wedge b = c $  for $ a,b, c \in B $ iff $I \wedge_{i\typearrow i\typearrow i } a b = c$.
		\item
		$ a = \neg b  $ for $ a,b \in B $ iff $I \neg_{i \typearrow i} a = b$.
		\item The valuation on $\mathcal{B}$ is defined such that for all $p^j \in X$,  $V(p^j)= I(p_i^j)$.
		\item
		$ (a,b) \in N^V $ for $ a,b \in B $ iff $IN_{i\typearrow\proptype} ab = T$.
	\end{itemize}
	
	Since ${H} \models^\text{HOL} \Sigma $ for all
	$ \Sigma \in \{COM \vee,..., Dis \wedge \vee  \} $, it is
	straightforward (but tedious) to verify that $ \wedge$, $ \vee $,  $\neg$, $ 0 $ and $ 1 $ satisfy the 
	conditions required for a Boolean algebra.
	
	Moreover, the above construction ensures that $H$ is a Henkin
	model $H^\mathcal{N}$ for Boolean normative model  $\mathcal{N}$. Hence, Lemma~\ref{lemmaab:3} applies. This
	ensures that for all conditional norms $(\varphi,\psi) $, and for all assignment $g$,  we have:
	\begin{center}
	    $ \| \lfloor d_{i}(N) (\varphi,\psi) \rfloor \|^{{H}, g }$ $ = T$ if and only if 
	$V(\psi) \in out^{\mathcal{B}}_{i} (N^V, \{V(\varphi)\})  $.
	\end{center}
	
  \subsection*{Proof for Theorem \ref{th:soco}}	  
   
   \subsubsection*{Soundness}
   \sloppy The proof is by contraposition. Suppose that for a Boolean normative model $  \langle \mathcal{B}, V ,N^V\rangle $, we have $V(\psi) \notin out^{\mathcal{B}}_{i} (N^V, \{V(\psi)\})  $. Now let $H^\mathcal{N}$ be a Henkin model for Boolean normative model $\mathcal{N}$. Then by Lemma~\ref{lemmaab:3} for an arbitrary assignment $g$, it is held that $\| \lfloor d_{i}(N)(\varphi,\psi) \rfloor \|^{{H}^{\mathcal{N}}, g } = F$, but $\| COM \vee    \|^{{H}^{\mathcal{N}}, g } = T$, ..., $\|    Dis \wedge \vee  \|^{{H}^{\mathcal{N}}, g } = T$, and that is a contradiction. 
	
	\subsubsection*{Completeness}
   The proof is again by contraposition. If it is assumed that $ \{COM \vee,..., Dis \wedge \vee  \} \nvDash^\text{HOL}
	\lfloor d_{i}(N)(\varphi,\psi) \rfloor  $, then there is a Henkin model ${H}=\langle \{D_\alpha\}_{\alpha \in {T}}, I \rangle$ such that $H\models^\text{HOL}$~$\Sigma $ for all $\Sigma \in \{ COM \vee,..., Dis \wedge \vee \}$, but
	$\| \lfloor d_{i}(N)(\varphi,\psi) \rfloor \|^{{H},g} = F$
	for some assignment $g$. By Lemma~\ref{lemmaab4}, there is a Boolean normative model $\mathcal{N}$ such that $ V(\psi) \notin out^{\mathcal{B}}_{i} (N^V, \{V(\varphi)\}) $, and that is a contradiction.


%\begin{thebibliography}{00}


%\bibitem{}

%\end{thebibliography}
%%%%%%%%%%
%\bibliography{Bibliography}
%%%%%%%%%%%
\end{document}